\documentclass{article}
\usepackage{acflow_preprint,times}

\usepackage{amsmath,amsfonts,bm}

\def\eqref#1{equation~\ref{#1}}

\def\1{\bm{1}}

\DeclareMathAlphabet{\mathsfit}{\encodingdefault}{\sfdefault}{m}{sl}
\SetMathAlphabet{\mathsfit}{bold}{\encodingdefault}{\sfdefault}{bx}{n}

\newcommand{\R}{\mathbb{R}}

\usepackage{hyperref}
\hypersetup{
  pdftitle={AcFlow: Controlling Text-to-Image Diffusion Transformers via Learned Conditional Activation Flow},
  pdfauthor={Junran Wang, Zehao Jin, Tianyu Luan, Ruixuan Deng, Tian Qiu, Xinjie Shen},
  pdfsubject={AcFlow preprint}
}
\usepackage{url}
\usepackage{amsmath}
\usepackage{amssymb}
\usepackage{booktabs}
\usepackage{graphicx}
\usepackage{multirow}
\usepackage{xcolor}
\usepackage{adjustbox}
\usepackage{makecell}
\usepackage{colortbl}
\usepackage{xspace}
\usepackage{wrapfig}
\usepackage[normalem]{ulem}
\usepackage{pifont}
\makeatletter
\newcommand{\captionof}[1]{\def\@captype{#1}\caption}
\makeatother

\definecolor{oursbg}{HTML}{E9E6FF}
\newcommand{\ours}{\textsc{AcFlow}\xspace}
\newcommand{\vphi}{v_{\phi}}
\newcommand{\cmark}{\ding{51}}
\newcommand{\xmark}{\ding{55}}

\title{\ours: Controlling Text-to-Image Diffusion Transformers via Learned Conditional Activation Flow}

\author{
Junran Wang\textsuperscript{*}, Zehao Jin\textsuperscript{*}, Tianyu Luan,\\
\bfseries Ruixuan Deng, Tian Qiu, Xinjie Shen\\[4pt]
\normalfont\textsuperscript{*}Equal contribution.
}

\begin{document}

\maketitle

\begin{abstract}
Text-to-image diffusion transformers (DiTs) are powerful generators, yet direct prompting provides limited control interface for style intensity and can fail to suppress unwanted concepts.
To enable these controls, we introduce \ours, an inference-time controller that transports intermediate layer image-token activations through a learned concept-conditioned velocity field while keeping the base DiT frozen.
A textual concept description specifies the desired intervention, while the integration horizon provides a continuous control parameter.
The field produces token-varying, activation-dependent updates.
With parameters shared across concepts within each task family, one field covers over 15,000 style descriptions or over 1,000 suppression concepts, and generalizes to concepts unseen during training without per-concept fitting.
On style control, \ours achieves the best style--content trade-off among the evaluated baselines in the high-style-alignment regime.
At a fixed operating point, \ours attains style--content alignment of 0.5365/0.2860, compared with 0.4397/0.2684 for the baseline with the highest style alignment.
On concept suppression, \ours reduces the fraction of images showing the concept from 95.3\%/82.1\% to 41.6\%/40.5\% on held-in/held-out concepts, including cases where deleting them from the prompt fails to remove them.
Our analyses support the learned velocity field as an adaptive control mechanism, with update directions varying across tokens and depending on their activation states.
Our code is available at \url{https://github.com/Nove1yst/AcFlow}.
\end{abstract}

\section{Introduction}
\label{sec:intro}

Text-to-image (T2I) diffusion models~\citep{rombach2022high} have become powerful tools for visual creation, yet their generation capabilities do not translate into precise user control.
A model may readily generate a stylized image without allowing users to smoothly adjust the degree of stylization, or depict an unwanted concept implied by the the prompt even when requested to exclude it.
Verbal modifiers such as 'weak' or 'strong' produce similarly pronounced level of stylization, and appending instructions such as 'no lighthouse' can leave the unwanted concept visibly present~\citep{rodriguez2025controlling}.

Existing control mechanisms offer limited flexibility for these requirements.
Parameter-level adaptation can provide a control interface, but methods based on concept-specific optimization require additional fitting for each new concept~\citep{chiu2026text,baumann2025continuous,gandikota2024concept}, affect non-target concepts~\citep{lyu2024one} or risk degradations of previously learned control capabilities during sequential modification~\citep{smith2023continual}.
This motivates an inference-time controller that shares parameters across concepts, acts upon a user-specified concept with an explicit parameter for controlling the intervention strength, while the generator's parameters remain frozen.

Intermediate activations offer a promising interface:
FluxSpace~\citep{dalva2024fluxspace} demonstrates semantic control through intermediate representations.
However, DiTs lack U-Net's explicit multiscale hierarchy and jointly update text and image states~\citep{rombach2022high}, complicating intervention site selection and the prediction of downstream effects from local activation interventions.
This motivates learning activation-dependent updates through the downstream generator.

\begin{figure}[h]
    \centering
            \caption{\textbf{Overview of \ours.}
\textbf{Upper: Method.} At one site in a frozen DiT, image-token states move through a velocity field conditioned on a concept description. Velocities vary with activation state across tokens. The displacement-direction map removes the rank-one PCA component: each cell is an image token, with color encoding its direction.
\textbf{Lower: Capabilities.} One field supports fine-grained descriptions and continuous control of styles. A separately trained suppression field removes the daisies concept. Held-out means concepts are unseen during training.}
    \includegraphics[width=0.9\linewidth]{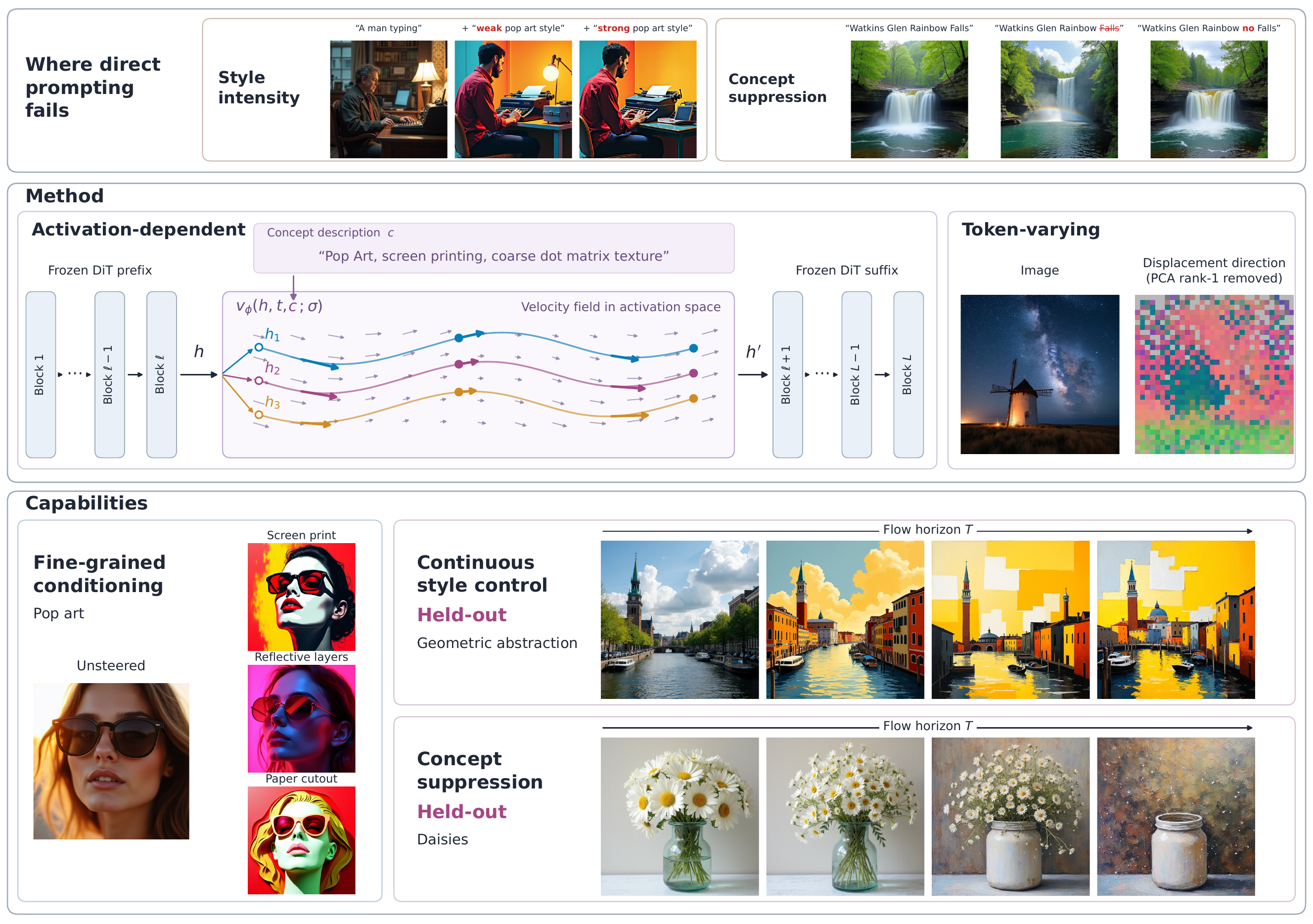}
    \label{fig:acflow_overview}
\end{figure}

To this end, we introduce \ours, which controls DiTs by transporting intermediate layer activations through a learned concept-conditioned velocity field at inference time.
A textual concept description specifies the desired intervention, and the integration horizon $T$ provides a continuous control parameter for the extent of intervention, with $T=0$ recovering the unmodified generator.
At a single intervention site, image-token activations moves through the field with velocities that adapt to their individual states and the concept condition, yielding token-varying activation-dependent updates.
We train the field through velocity distillation, sharing its parameters across concepts within each task family, e.g. style control or concept suppression. The field supports fine-grained descriptions for style control and generalizes to unseen concepts during training without concept-specific fitting.
Our main contributions are:

\textbf{Control as a flow field.} We formulate inference time control as transport through a learned velocity field in activation space. Image-token activations move through the field, receiving token-varying, activation-dependent updates.

\textbf{Concept-shared control and generalization.} Conditioned on textual concept descriptions, one field is trained for one task family without per-concept fitting. The shared field covers over 15,000 style descriptions or over 1,000 suppression concepts, generalizes to unseen concepts during training and responds to fine-grained descriptions for style control beyond coarse labels.

\textbf{Continuous style control and concept suppression.} \ours enables progressive style modulation with the flow horizon as a continuous control parameter, achieving the best style--content trade-off among evaluated methods in the high-style-alignment regime. Concept suppression results further demonstrate the applicability where direct prompting leaves unwanted concepts present.

\textbf{Probing the learned field.} Controlled experiments demonstrate the value of token-specific updates, while trajectory analysis reveals that the field adapts its direction and energy allocation systematically through denoising steps. These findings support the value of an adaptive activation transport field beyond a fixed concept direction broadcast across tokens.

\section{Related Works}
\label{sec:related}

\textbf{Continuous concept control.}
Attribute Control~\citep{baumann2025continuous} identifies concept-specific directions in text embeddings through optimization, enabling continuous modulation.
Concept Sliders~\citep{gandikota2024concept} learns concept-specific low-rank adapters~\citep{hu2021lora} in the diffusion backbone and continuously adjusts their influence through an inference-time scale.
Text Slider~\citep{chiu2026text} learns such adapters in the text encoder, enabling continuous attribute control.
All approaches fit adapters for individual target concepts.
\ours instead trains a shared, text-conditioned field within each task family and uses the flow horizon to adjust intervention strength, including for concepts unseen during training.

\textbf{Activation intervention for T2Is.}
Recent methods have proposed to control generation by modifying a model's intermediate representations at inference time.
ICM~\citep{zaleska2026attention} uses linear probes to identify fixed attribute directions, while Semantic Steering~\citep{li2026semantic} constructs a fixed target-to-safe direction.
CASteer~\citep{gaintseva2026casteer} and SHIFT~\citep{konovalova2026shift} retain concept-specific directions but dynamically vary their magnitudes.
DSAS~\citep{ferrando2025dsas} learns activation-dependent scalar gates for a prescribed steering transformation.
PID Steering~\citep{nguyen2026activation} uses feedback to adjust the magnitude of a mean direction.
In a parallel branch, AcT~\citep{rodriguez2025controlling} estimates coordinate-wise affine maps between source and target activations.
SteeringDiffusion~\citep{wu2026steeringdiffusion} predicts condition-dependent affine modulation for multiple styles.
Dynamic scaling and affine maps extend fixed vectors while constraining updates to a prescribed operation family.

\textbf{Nonlinear activation intervention for T2Is.}
Nonlinear intervention allows the update direction to vary with the activation state.
Conditioned Activation Transport (CAT)~\citep{chrabaszcz2026conditioned} learns a geometry-gated nonlinear displacement, but computes it from a mean-pooled activation and broadcasts the same update across tokens.
FLAS~\citep{jin2026beyond} provides a flow formulation of steering intermediate representations of Large Language Models; \ours uses a learned concept-conditioned field to transport image-token activations.

\section{Method}
\label{sec:method}

\begin{table}[h!]
\caption{\textbf{Control interfaces and update geometry of T2I activation intervention.}
\textcolor{red}{\checkmark}: supported; $\times$: not established in the cited
T2I formulation. Concept conditioning excludes concept-specific fitting; zero-shot concerns unseen concepts during training.
Activation-dependent direction means the displacement axis can vary with $h$
at a fixed concept, site, and denoising step.}
\label{tab:steering_taxonomy}
\centering
\small
\setlength{\tabcolsep}{5.0pt}
\renewcommand{\arraystretch}{1.18}
\resizebox{\textwidth}{!}{\begin{tabular}{@{}llcccc@{}}
\toprule
\textsc{Method} & \textsc{Update family} &
\makecell{\textsc{Activation-dependent}\\\textsc{direction}} &
\makecell{\textsc{Token-varying}\\\textsc{direction}} &
\makecell{\textsc{Concept-conditioned}\\\textsc{controller}} &
\makecell{\textsc{Zero-shot}\\\textsc{concepts}} \\
\midrule
Linear-AcT~\citep{rodriguez2025controlling} & Affine map
    & \textcolor{red}{\checkmark} & \textcolor{red}{\checkmark} & $\times$ & $\times$ \\
DSAS (CAA)~\citep{ferrando2025dsas} & Gated direction
    & $\times$ & $\times$ & $\times$ & $\times$ \\
CAT~\citep{chrabaszcz2026conditioned} & Nonlinear map
    & \textcolor{red}{\checkmark} & $\times$ & $\times$ & $\times$ \\
SteeringDiffusion~\citep{wu2026steeringdiffusion} & Affine modulation
    & \textcolor{red}{\checkmark} & \textcolor{red}{\checkmark} & \textcolor{red}{\checkmark} & $\times$ \\
CASteer~\citep{gaintseva2026casteer} & Scaled direction
    & $\times$ & $\times$ & $\times$ & $\times$ \\
ICM~\citep{zaleska2026attention} & Fixed direction
    & $\times$ & $\times$ & $\times$ & $\times$ \\
SHIFT~\citep{konovalova2026shift} & Scaled direction
    & $\times$ & \textcolor{red}{\checkmark} & $\times$ & $\times$ \\
Semantic Steering~\citep{li2026semantic} & Fixed direction
    & $\times$ & $\times$ & $\times$ & $\times$ \\
\midrule
\rowcolor{oursbg}\ours & Velocity field
    & \textcolor{red}{\checkmark} & \textcolor{red}{\checkmark} & \textcolor{red}{\checkmark} & \textcolor{red}{\checkmark} \\
\bottomrule
\end{tabular}}
\end{table}

\subsection{Control as a Flow Field}
\label{sec:steering}

Let $G_{\theta}$ denote the frozen denoising velocity predictor of a DiT-based~\citep{peebles2023scalable} T2I model with hidden dimension $d$.
For one sample, the output of a selected single-stream block $b$ is a residual-stream state $h\in\R^{S\times d}$; we omit the batch dimension throughout.
Here $S=S_{\mathrm{txt}}+S_{\mathrm{img}}$ counts text and image tokens, and $h_i\in\R^d$ denotes token $i$.
Given a source prompt $p$ (written $p_{\mathrm{src}}$ in training) and a concept description $c$, we seek an edited state $h'$ that induces the requested concept change while preserving non-target content.
For example, for style control, $p$ is a content prompt and $c$ describes the style of the image (e.g., in pop art style, use reflective layers).

At a fixed denoising noise level $\sigma\in[0,1]$, we define an activation flow field
$\varphi_t(\cdot;c,\sigma):\R^{S\times d}\rightarrow\R^{S\times d}$, parameterized by the field weights $\phi$.
For brevity, we write $\varphi_t(h)$ with $c$, $\sigma$, and $\phi$ implicit.
The flow time $t\in[0,T]$ is distinct from $\sigma$: $c$ and $\sigma$ remain fixed during each activation-flow integration, which restarts from the current block output at every denoising step.

The velocity field $\vphi(h,t,c;\sigma)\in\R^{S\times d}$ generates the flow through
\begin{equation}
    \frac{d}{dt}\varphi_t(h) = \vphi(\varphi_t(h),t,c;\sigma), \qquad \varphi_0(h)=h.
\end{equation}
The steered activation at strength $T$ is
\begin{equation}
    h' = \varphi_T(h) = h + \int_0^T \vphi(\varphi_t(h),t,c;\sigma)\,dt.
\end{equation}
In practice, we use $N$ forward Euler steps:
\begin{equation}
    h^{(k+1)} = h^{(k)} + \frac{T}{N}\vphi\!\left(h^{(k)},\frac{kT}{N},c;\sigma\right),
    \quad k=0,\ldots,N-1,
\end{equation}
Here $N$ is the number of integration steps.
Starting from $h^{(0)}=h$, we write the numerical endpoint $h^{(N)}\approx\varphi_T(h)$ back into the residual stream.
The horizon $T$ controls intervention strength; $T=0$ leaves $h$ unchanged.
Several simpler methods arise by restricting $v_{\phi}(h,t,c)$:
\begin{list}{\textbullet}{    \setlength{\leftmargin}{1.5em}    \setlength{\labelsep}{0.5em}    \setlength{\topsep}{0pt}    \setlength{\partopsep}{0pt}    \setlength{\itemsep}{0pt}    \setlength{\parsep}{0pt}}
    \item $v_{\phi}(h,t,c)=g(\bar{h},t,c)$ where $\bar{h}$ is the mean-pooled token applies a shared velocity to all tokens;
    \item $v_{\phi}(h,t,c)=a(h,t,c)d(c)$ where $a$ is a scalar fixes a concept-specific direction with scaling;
    \item $v_{\phi}(h,t,c)=d(c)$ reduces the flow to $\varphi_T(h)_i=h_i+T d(c)$, the standard additive intervention.
\end{list}
The full field relaxes both the shared-token update and fixed concept direction restrictions: tokens move through activation space with velocities that can change in direction as well as magnitude.

\subsection{Intervention Site and Token Selection}
\label{sec:image_tokens}

Prior work demonstrates that intermediate DiT representations support semantic control~\citep{dalva2024fluxspace}, while analyses reveal depth-dependent contributions of single-stream blocks to content and style formation~\citep{yang2026splitflux}.
Following the use of intermediate layers in LLM activation steering~\citep{turner2024activation,rimsky2024steering}, we choose one block near the middle of the single-stream stack, where semantic representations have been established by the preceding blocks, while the downstream layers integrate the intervention into the denoising prediction.

Write the single-stream state as
$h=[h^{\mathrm{txt}};h^{\mathrm{img}}]$, where
$h^{\mathrm{txt}}\in\R^{S_{\mathrm{txt}}\times d}$,
$h^{\mathrm{img}}\in\R^{S_{\mathrm{img}}\times d}$, and $[\cdot;\cdot]$ denotes concatenation along the token dimension.
Although DiTs typically mix the two token groups in their single-stream blocks,
their roles at the intervention site are asymmetric.  Intervention
$h^{\mathrm{txt}}$ can affect the output only indirectly through
later text--image mixing, whereas $h^{\mathrm{img}}$ lies on the residual path
consumed by the output head.  
We consequently intervene on only the image state:
\begin{equation}
    h' = [h^{\mathrm{txt}};
          \varphi_T^{\mathrm{img}}(h^{\mathrm{img}})],
    \qquad
    \varphi_t^{\mathrm{img}}:
    \R^{S_{\mathrm{img}}\times d}\rightarrow
    \R^{S_{\mathrm{img}}\times d}.
    \label{eq:image_only_flow}
\end{equation}

\begin{wrapfigure}{r}{0.5\textwidth}
    \centering
        
    \includegraphics[width=\linewidth]{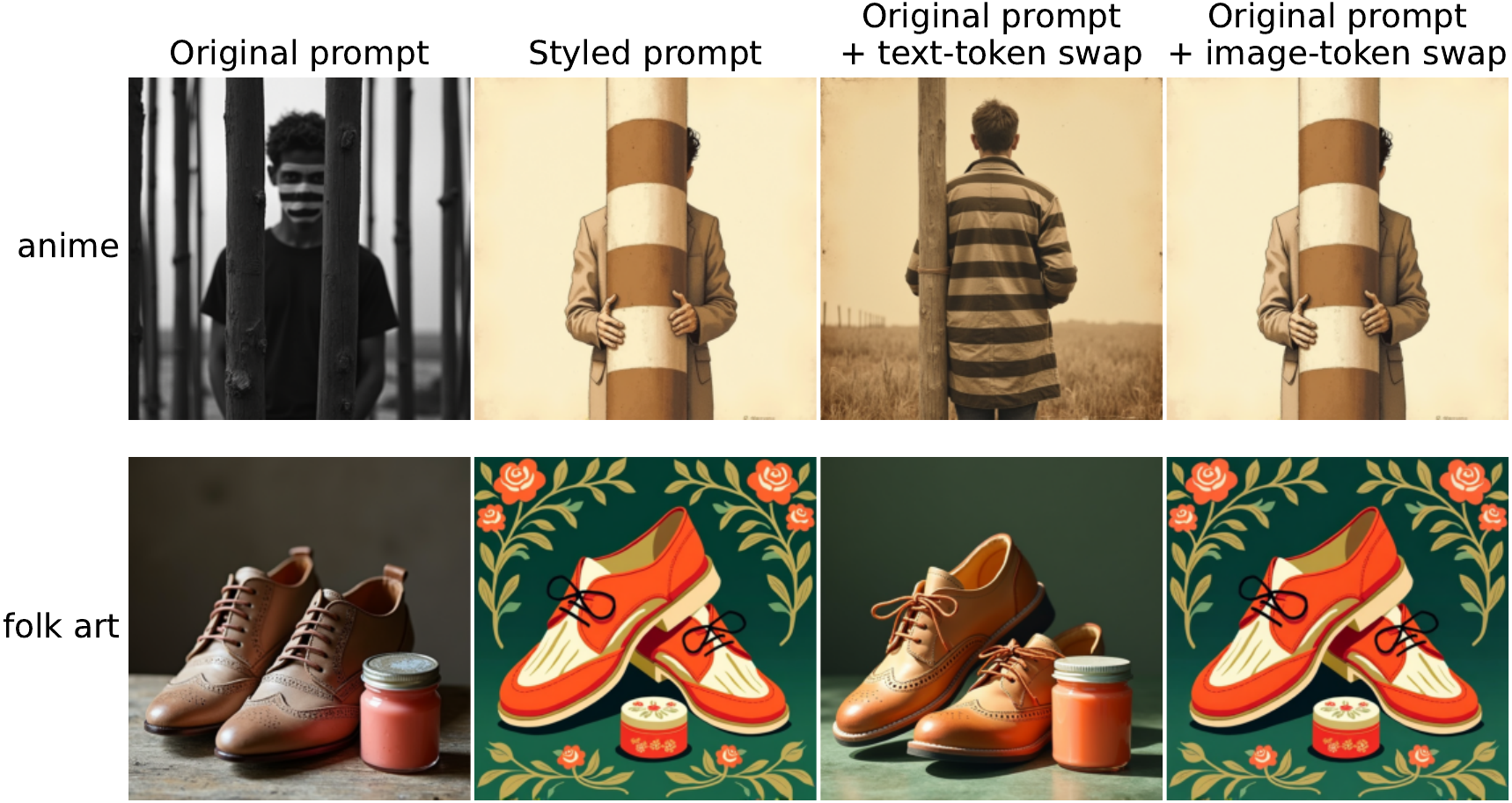}
    \caption{\textbf{Token-swap:}
    A target-prompt trajectory supplies block outputs to a source-prompt
    trajectory with the same initial noise.}
    \label{fig:image_token_oracle}
    
\end{wrapfigure}

Below, $\vphi$ denotes its image-token component.
Figure~\ref{fig:image_token_oracle} shows the diagnostic used to check this choice.
At every denoising step, we record the output of single-stream block 16 of FLUX.1-dev~\citep{flux1dev2024} under a target prompt with style description and replace the text or image tokens in a same noise source trajectory under the original prompt without the style description.
The diagnostic shows that image-state replacement shows significant stylization and recovers 93.4\% style alignment score, while text-state replacement recovers only 46.1\%; full results appear in App.~\ref{app:intervention_details}.

\subsection{Model Architecture}
\label{sec:architecture}

We instantiate $\vphi$ with one backbone-matched Transformer-style FlowBlock.
The condition follows the generator's native text path:
let $C=E_{\mathrm{txt}}(c)\in\R^{S_c\times d}$ be text-encoder features and $\bar c=E_{\mathrm{pool}}(c)$ be a separate pooled text-encoder feature when $G_{\theta}$ consumes pooled feature.

The FlowBlock first extracts concept information through cross-attention.  Let
$e_{\mathrm{flow}}(t)$ embed the current integration time, $e_{\mathrm{model}}(\sigma)$ the noise level and $e_c(\bar c)$
project the pooled condition. Following the backbone's adaptive normalization and residual gate,
\begin{align}
    x_t,g_t &={\rm AdaNorm}_{\phi}
        \!\left(h^{\mathrm{img}};e_{\mathrm{flow}}(t)+e_c(\bar c)\right),\\
    \tilde h & = h^{\mathrm{img}} + g_t \odot
        {\rm CrossAttn}_{\phi}(Q=x_t,K=C,V=C).
\end{align}
We preserve the backbone's attention geometry, QK normalization,
normalization placement, and gated-residual convention.  The sole change to
its time modulation is to replace the generator's model time by flow time $t$,
allowing the concept update to vary along the integration trajectory.

The cross-attention stage is followed by one native-style transformer block,
\begin{equation}
    h^{\mathrm{out}} = {\rm TransformerBlock}_{\phi}
        \!\left(\tilde h;e_{\mathrm{model}}(\sigma)+e_c(\bar c)\right),
    \qquad
    \vphi(h^{\mathrm{img}},t,c;\sigma)=h^{\mathrm{out}}-h^{\mathrm{img}}.
\end{equation}

\subsection{Velocity Distillation}
\label{sec:distillation}

Given training samples $(z_0,p_{\mathrm{src}},p_{\mathrm{tgt}},c)\sim\mathcal D$, where $z_0$ is the encoded target image, the prompt pair specifies the concept change $c$.
We sample $\epsilon\sim\mathcal N(0,I)$ with the same shape as $z_0$ and $\sigma\sim q_{\sigma}$ on $[0,1]$, where $q_{\sigma}$ is the training noise-level distribution, and construct
\begin{equation}
    z_{\sigma} = (1-\sigma)z_0 + \sigma\epsilon.
\end{equation}
The frozen teacher is the unsteered denoising predictor conditioned on the target prompt:
\begin{equation}
    u_{\mathrm{tgt}} = G_{\theta}(z_{\sigma}, \sigma, p_{\mathrm{tgt}}).
\end{equation}
The student is the same frozen model conditioned on the source prompt, but with \ours active at block $b$ using condition $c$:
\begin{equation}
    \hat{u} = G_{\theta,\phi}(z_{\sigma}, \sigma, p_{\mathrm{src}}, c, T).
\end{equation}
Teacher and student use the same $z_{\sigma}$, $\sigma$, and guidance setting.
For training, we set $T=1$ and minimize the mean squared velocity error by optimizing $\phi$,
\begin{equation}
    \mathcal L_{\mathrm{vel}}(\phi)
    =\mathbb E_{\mathcal D,\epsilon,\sigma}\!\left[
    \operatorname{MSE}\!\left(\hat u,\mathrm{stopgrad}(u_{\mathrm{tgt}})\right)\right].
\end{equation}

\section{Experiments}
\label{sec:experiments}

We test the capability of \ours on two control tasks: continuous style control and concept suppression, both not reliably achievable by directly prompting DiTs.
Unless stated otherwise, experiments use frozen FLUX.1-dev at
$512\times512$ resolution, 28 denoising steps, guidance 3.5, single-stream block $b=16$ and train with $T=1$ and $N=3$. Z-Image~\citep{cai2025z} portability is in App.~\ref{app:z}; more applications are in App.~\ref{app:additional_applications} and App.~\ref{app:hub_ip}.

\subsection{Continuous Style Control}
\label{sec:megastyle}

\textbf{Fine-grained style conditions.}
MegaStyle~\citep{gao2026megastyle} specifies each image's style through a
fine-grained natural-language description paired with a content prompt. We group
these descriptions into \emph{style families} to
construct family-disjoint held-in and held-out splits, containing 39 and 6 distinct style families, respectively.
A single conditional field is trained on over 15,000 distinct style descriptions from the held-in style families, and none of the 704 held-out descriptions are seen during training.
At generation time, the generator receives the content prompt and the
field receives the style description.
For the metrics, style alignment is the reference-image cosine similarity from the released MegaStyle encoder; content alignment is the CLIP~\citep{radford2021learning} similarity
to the content prompt.

\textbf{Baselines.}
\label{sec:style_baselines}
We compare activation intervention methods: Linear-AcT~\citep{rodriguez2025controlling} at its original
locations and at our image-token site, the corresponding Mean-AcT
variants, a diffusion adaptation of Activation
Addition~\citep{turner2024activation},
SHIFT~\citep{konovalova2026shift} and parameter adaptation methods:
Concept Sliders~\citep{gandikota2024concept} and Text Slider~\citep{chiu2026text}.
For baseline comparisons, we fix one style description per held-in style family across methods.
While the baselines fit style-specific interventions, \ours uses a single field shared across style families. Implementation details appear in App.~\ref{app:repro_megastyle}.

\begin{figure}[ht]
    \centering
    \begin{minipage}[t]{0.49\textwidth}
        \vspace{0pt}
        \centering\small
\setlength{\tabcolsep}{1pt}
\renewcommand{\arraystretch}{1.05}
\begin{tabular}{@{}lcccc@{}}
\toprule
\multirow{2}{*}{\textbf{Method}} & \multicolumn{2}{c}{\textbf{Held-in}} & \multicolumn{2}{c}{\textbf{Held-out}} \\
\cmidrule(lr){2-3}\cmidrule(lr){4-5}
 & Style $\uparrow$ & Content $\uparrow$ & Style $\uparrow$ & Content $\uparrow$ \\
\midrule
Styled prompt & 0.548 & 0.294 & 0.511 & 0.287 \\
\midrule
\multicolumn{5}{@{}l}{\textbf{Parameter Adaptation}} \\
Concept Sliders & 0.511 & 0.279 & -- & -- \\
Text Slider & 0.157 & 0.306 & -- & -- \\
\midrule
\multicolumn{5}{@{}l}{\textbf{Activation Intervention}} \\
Linear-AcT & 0.271 & 0.202 & -- & -- \\
Linear-AcT (b16) & 0.313 & 0.176 & -- & -- \\
Mean-AcT & 0.440 & 0.268 & -- & -- \\
Mean-AcT (b16) & 0.298 & 0.277 & -- & -- \\
ActAdd (b16) & 0.300 & 0.278 & -- & -- \\
SHIFT & 0.221 & 0.279 & -- & -- \\
\midrule
\rowcolor{oursbg}\ours & \textbf{0.537} & \textbf{0.286} & \textbf{0.442} & \textbf{0.281} \\
\bottomrule
\end{tabular}

        \captionof{table}{\textbf{Style control,} one style description per style family.
        \ours is evaluated at $T=2$ on the same field; baselines are fitted per style and use their maximum-style operating points.}
        \label{tab:megastyle_operating_points}
    \end{minipage}\hfill
    \begin{minipage}[t]{0.49\textwidth}
        \vspace{0pt}
        \includegraphics[width=\linewidth]{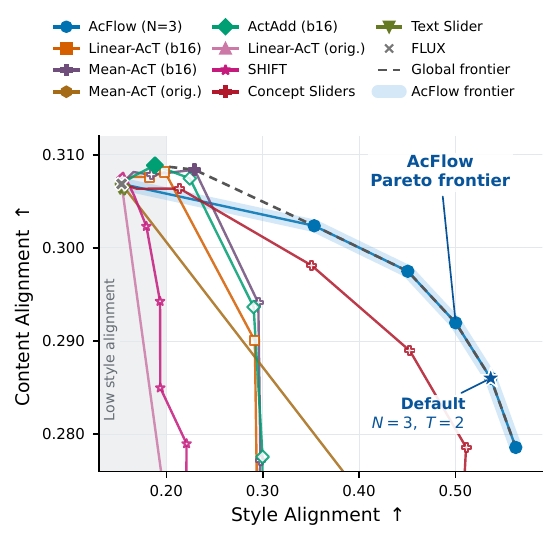}
        \caption{\textbf{Style--content trade-offs.}
        \ours is on the global frontier in high style-alignment region. 
        The complete range and
        uncertainty intervals appear in App.~\ref{app:complete_quantitative}.}
        \label{fig:megastyle_pareto}
    \end{minipage}
\end{figure}

Table~\ref{tab:megastyle_operating_points} compares methods at a fixed operating point. 
\ours is the strongest in terms of both style and content alignment among the displayed methods in the high-style-alignment region at their reported operating points, reaching 0.5365 and 0.2860, respectively, compared with 0.4397 and 0.2684 for
the strongest activation intervention method by style, Mean-AcT, and
the parameter adaptation baseline, Concept Sliders, which reaches 0.5109 and 0.2786.
Figure~\ref{fig:megastyle_pareto} complements these operating points by
comparing the full style--content trade-off. Among the evaluated methods, \ours offers the strongest style--content
trade-off in the high-style-alignment regime.

\begin{wrapfigure}{r}{0.55\textwidth}
        
    \centering
    \includegraphics[width=\linewidth]{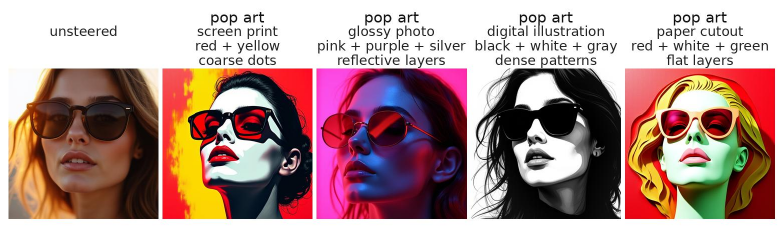}
    \caption{\textbf{Fine-grained conditioning.}
    Images share the same content prompt, style family and $T=2$ but style descriptions vary: screen-print halftones, glossy reflective photography, grayscale linear illustration, and layered paper cutout.}
    \label{fig:finegrained}
        
\end{wrapfigure}

\begin{figure}[t]
    \centering
    \includegraphics[width=0.9\linewidth]{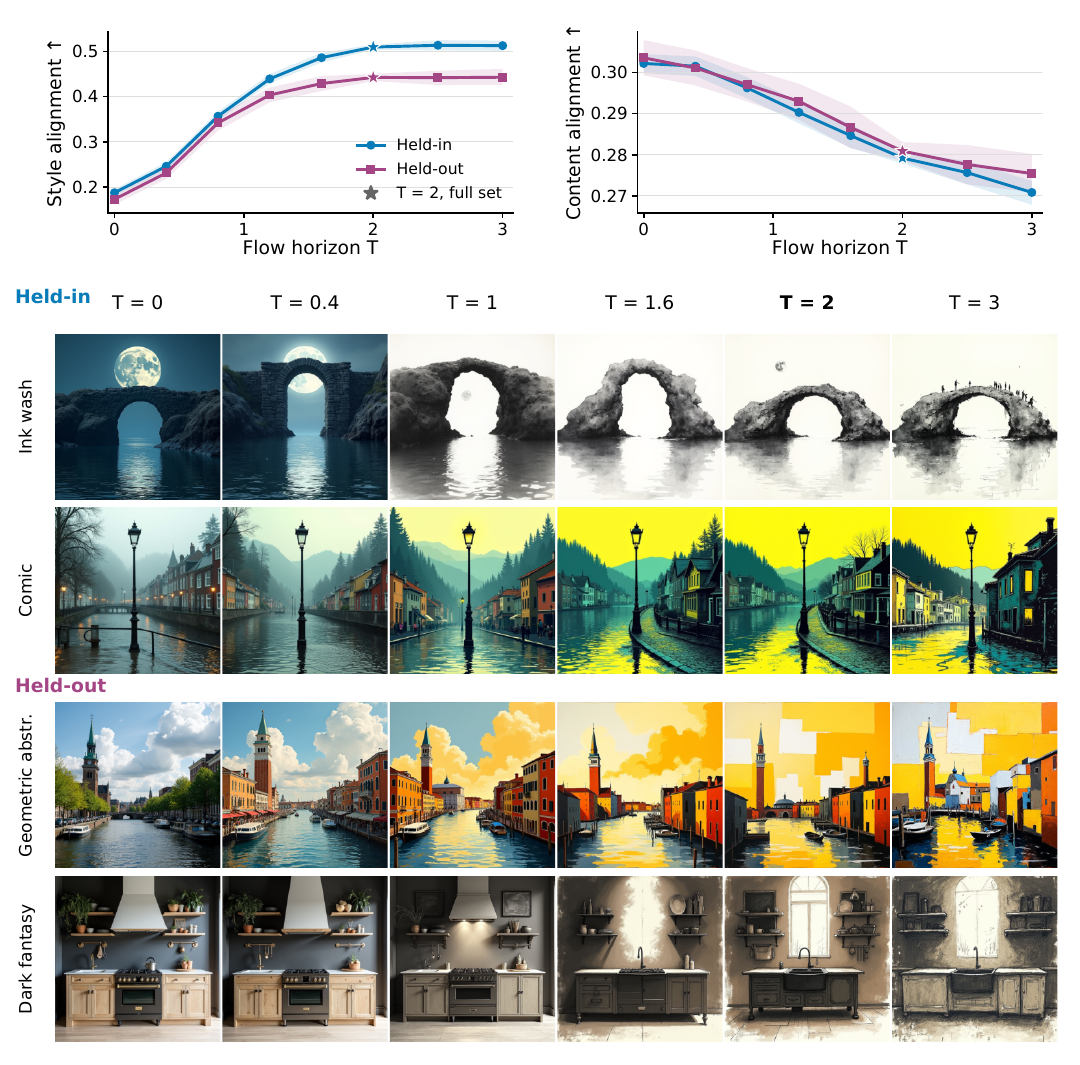}
    \caption{\textbf{Continuous style control and generalization.} \ours is evaluated using the original fine-grained style descriptions and paired content prompts.
    Top: \ours's strength response. $T=2$
    use full test sets; others use fixed quarter
    subsets. Bands show 95\% within-family bootstrap intervals. Bottom: illustrative sequences with fixed content and style. Increasing $T$ strengthens
    stylization, while large horizons can also alter composition.
    More sequences appear in App.~\ref{app:supplementary_style}.}
    \label{fig:megastyle_strength}
\end{figure}

\textbf{One field for fine-grained control within and across style families.}
To further demonstrate \ours's capability, we separately evaluate \ours using the original fine-grained style descriptions and paired content prompts on both held-in and held-out style families.
Fig.~\ref{fig:finegrained} demonstrates control over distinct visual treatments within one style family.
Across families, Fig.~\ref{fig:megastyle_strength} shows that the same field supports different style concepts without additional fitting.

\textbf{Continuous style control.}
Flow horizon $T$ provides an inference-time coordinate along the style--content trade-off.
Figure~\ref{fig:megastyle_strength} shows increasing style alignment over the
responsive range on both held-in and held-out style families before saturation, accompanied by a
gradual reduction in content alignment. The fixed-noise sequences illustrate
the corresponding gradual changes in style.
Although the field is trained at $T=1$, extending the horizon further
strengthens stylization on both seen and unseen families, extending the useful control range.

\textbf{Generalization to unseen style families.}
To test transfer beyond new descriptions of familiar styles, we test on style families excluded from training. On all unseen families, \ours reaches $0.442$ style alignment and $0.281$ content alignment at
$T=2$. Fig.~\ref{fig:megastyle_strength} shows that \ours is capable of inducing visible stylization as well as controlling the strength of stylization on held-out families.

\subsection{Concept Suppression}
\label{sec:concept_negation}
We test the same formulation on concept suppression. From removal pairs mined from InstructPix2Pix~\citep{brooks2023instructpix2pix}, we build concept-disjoint
splits. A single field is trained on over 1,000 held-in concepts and evaluated on 150 held-out concepts that never appear in training.
The source prompt contains the unwanted concept, the teacher uses the edited prompt, and the
field receives \texttt{erase <concept>}. Implementation details are in App.~\ref{app:repro_suppression}.
  
\textbf{One field for held-in and held-out concepts.} We generate images at $T\in\{0,1,2,3\}$ from the same initial noise and ask InternVL2.5-8B-MPO~\citep{chen2024expanding} whether the suppressed concept is clearly visible. 
The fraction of images showing the concept falls monotonically with $T$, from 95.3\%/82.1\% at $T{=}0$ to 41.6\%/40.5\% at $T{=}2$ on held-in/held-out concepts, so suppression transfers to held-out concepts.  The cost at $T=2$ is small: CLIP similarity to the concept-free prompt drops by about 0.015 and quality metric MUSIQ~\citep{ke2021musiq} by about 2 points.

We note that these prompts are demanding as the concept-free prompt still implies the concept, e.g. \emph{"snow"} in \emph{"Picture winter, street, home"}, and generating from the edited prompt leaves the concept visible in 28.9\%/15.6\% of images. On those cases where prompt editing fails, \ours suppresses the concept in 60.5\%/73.9\% at $T{=}3$. More examples, analyses and the full table are in App.~\ref{app:ip2p_quantitative}.

\begin{figure}[t]
    \centering
    \includegraphics[width=0.9\linewidth]{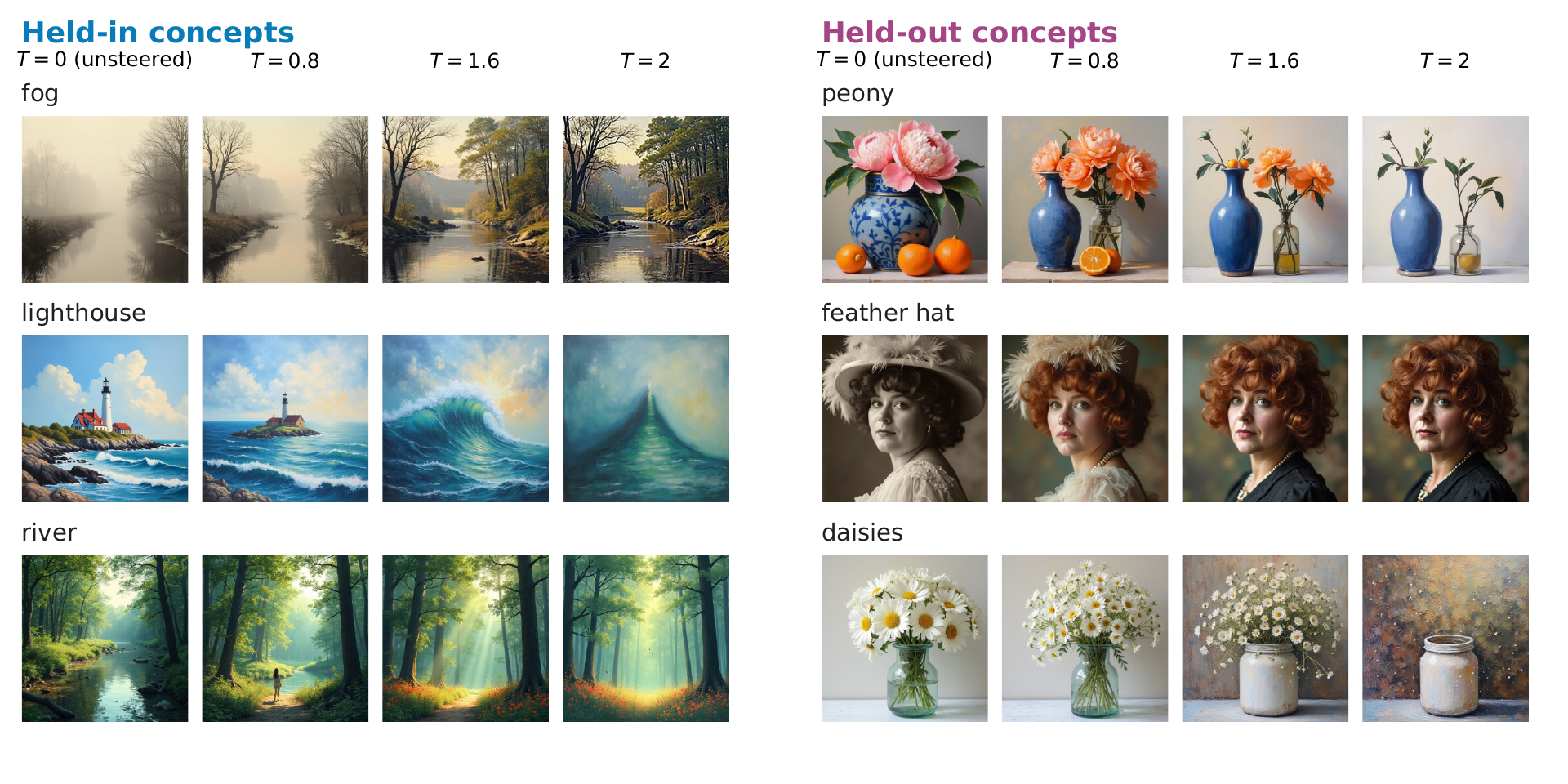}
    \caption{\textbf{Concept suppression examples.}
    Left: three held-in concepts; right: three held-out concepts. Each row fixes
    the original prompt and initial noise and varies $T$ using the same field. More sequences appear in App.~\ref{app:supplementary_negation}.}
    \label{fig:ip2p_strength}
\end{figure}

\textbf{Progressive suppression.} Fig.~\ref{fig:ip2p_strength} illustrates \ours's capability of conditional concept suppression on both held-in and held-out concepts.
Intermediate horizons expose partial suppression: the lighthouse becomes
smaller, the river narrows, and peony blooms give way to sparse
stems as $T$ increases. Recognizable elements remain, although
appearance and composition can also change.

\section{Ablations}
\label{sec:ablations}

We ablate the two main design choices: number of Euler steps $N$ and intervention location $b$ on style control. The generator is FLUX.1-dev at $512\times512$ resolution, 28 denoising steps and guidance 3.5. For each setup, we train a separate field under the same training budget. We retain $T=1$ as training horizon and evaluate at $T=2$. For evaluation, we use the original fine-grained style description and content.
Table~\ref{tab:style_ablation} summarizes the results. The complete flow horizon responses apppear in App.~\ref{app:complete_quantitative}.

\begin{table}[tbp]
\centering
\begin{minipage}{0.6\textwidth}
 
\centering\footnotesize
\setlength{\tabcolsep}{2pt}
\begin{tabular*}{\linewidth}{@{\extracolsep{\fill}}lrrrrr@{}}
\toprule
& & \multicolumn{2}{c}{Held-in} & \multicolumn{2}{c}{Held-out} \\
\cmidrule(lr){3-4}\cmidrule(l){5-6}
$b$ & $N$ & Style $\uparrow$ & Content $\uparrow$ & Style $\uparrow$ & Content $\uparrow$ \\
\midrule
\multicolumn{6}{@{}l}{\textbf{Intervention block}} \\
12 & 3 & 0.541 & 0.279 & 0.475 & 0.281 \\
20 & 3 & 0.508 & 0.282 & 0.464 & 0.282 \\
\midrule
\multicolumn{6}{@{}l}{\textbf{Euler steps}} \\
16 & 1 & 0.432 & 0.284 & 0.375 & 0.285 \\
16 & 3 & 0.510 & 0.279 & 0.442 & 0.281 \\
16 & 5 & 0.485 & 0.280 & 0.433 & 0.281 \\
16 & 10 & 0.493 & 0.283 & 0.439 & 0.286 \\
\bottomrule
\end{tabular*}
 
\captionof{table}{Style control ablations at $T=2$.}
\label{tab:style_ablation}
\label{tab:location_ablation}
\label{tab:euler_ablation}
 
\end{minipage}
\end{table}

\textbf{Intervention block.}
\ours maintains strong across blocks 12, 16,
and 20, with similar content scores on both splits. This
supports robustness to block selection within the tested range: effective
intervention does not require a specially chosen site.
Block 12 gives the highest mean style scores (0.541/0.475 for held-in/held-out),
while block 20 remains comparable to block 16 on held-in styles
and improves the held-out mean from 0.442 to 0.464. Location therefore
provides room for further improvement without being critical to the effectiveness.

\textbf{Euler integration steps.}
At block 16, Euler-3 gives the highest mean style scores on both splits.
Decreasing the count to one draws back style alignment significantly, while
increasing the count to five or ten does not improve style scores
monotonically; Euler-10 instead gives the highest content scores among configurations with competitive style alignment.
These results support Euler-3 as a compact configuration for strong style control.

\section{Characterizing the Flow Field}
\label{sec:analysis}
\label{sec:token}

We examine how the learned field adapts its updates across image tokens
and denoising noise levels. All experiments in this section are done with the default $N=3, b=16$ field evaulated at $T=2$.

\subsection{Variation Across Image Tokens}
\label{sec:shared_direction}
\label{sec:direction_effect}

For a fixed sample, concept, and denoising step, let
$\Delta\in\mathbb{R}^{S_{\mathrm{img}}\times d}$ stack the endpoint
displacements. A shared-axis update $\Delta_i=a_i u$ permits signed,
token-dependent magnitudes but has rank at most one. With $u_1$ the leading
right-singular vector from uncentered PCA, decompose
\begin{equation}
    q_i=\langle\Delta_i,u_1\rangle u_1,\qquad
    r_i=\Delta_i-q_i,\qquad \Delta=Q_1+R.
    \label{eq:token_rank1}
\end{equation}
We compare the full field with $Q_1$ rescaled to $\|\Delta\|_F$ and with
$R$ at its original norm, recomputing both the field and axis on each trajectory at every denoising step.

\begin{figure}[h]
    \centering
    \includegraphics[width=0.8\linewidth]{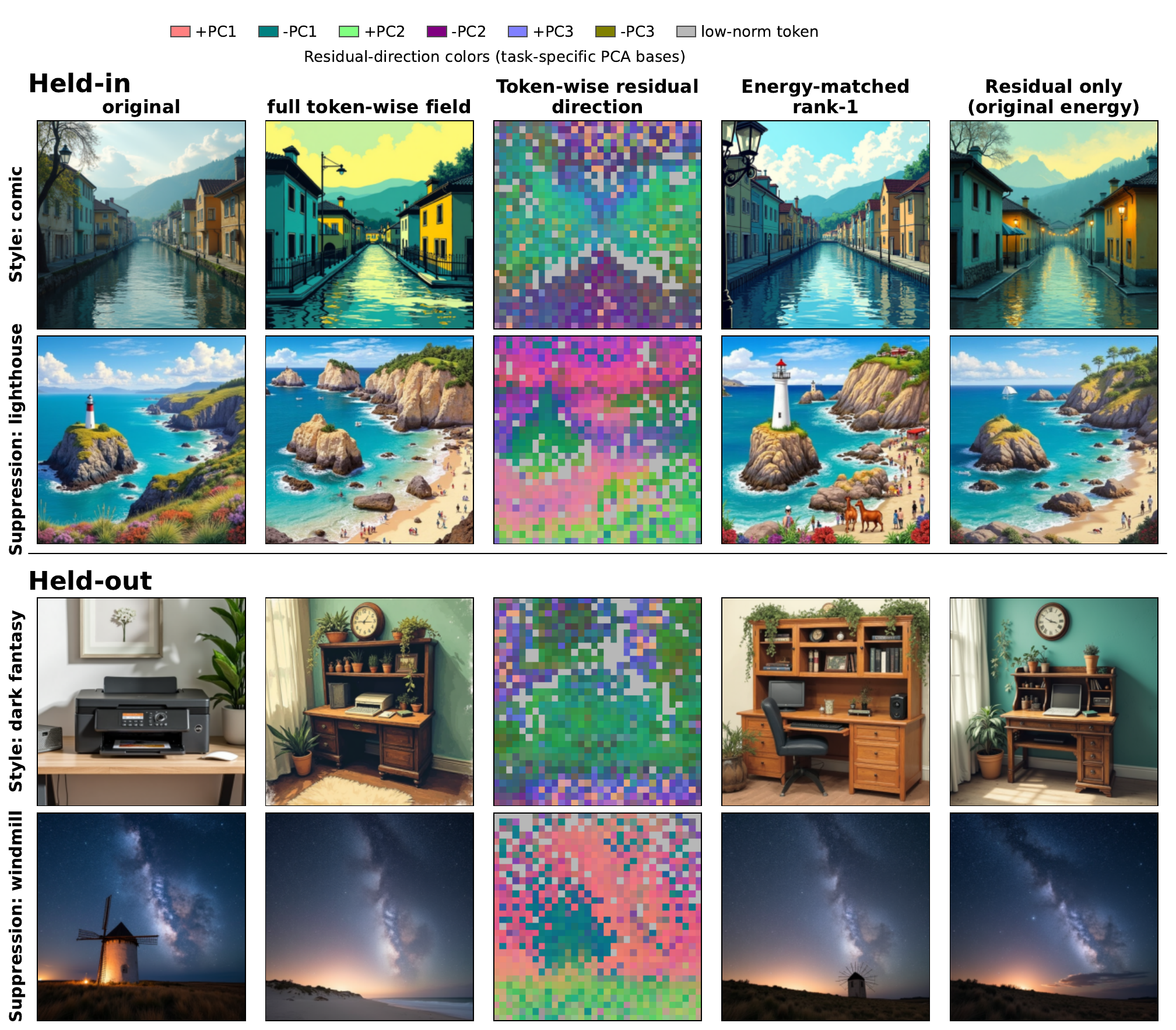}
    \caption{\textbf{Token-direction controls.} Selected examples compare full field, norm-matched rank-one update, and original-norm residual. Residual-direction colors use a three-component
    PCA basis shared within each case, and each cell corresponds to an image token.}
    \label{fig:token_direction_analysis}
\end{figure}

\textbf{The contribution of directions beyond the dominant shared axis.}
$Q_1$ carries most of the displacement energy: 88.7\%/90.7\% on held-in/held-out suppression
concepts. Yet restricting the update to $Q_1$ and matching its norm to the
full update recovers only
15.6\%/29.1\% of the full-field suppression effect, where suppression is
judged by the VLM of Sec.~\ref{sec:concept_negation}. The residual at its
original norm retains
26.0\%/50.6\% of the suppression effect; rescaled to the full norm, it
retains 100.0\%/108.9\% at the same image quality and content alignment as the full field.
Style control shows a similar pattern.
These evidence support the contribution of directions beyond the dominant shared axis with the largest energy. Fig.~\ref{fig:token_direction_analysis} illustrates residual-only effects and spatially coherent residual directions.
App.~\ref{app:per_token_geometry} reports the complete
comparison.

\subsection{Variation Across Noise Levels}
\label{sec:noise_geometry}
We next apply the same decomposition to each full-field denoising trajectory
and measure the off-axis energy fraction
$\rho(\sigma)=\|R(\sigma)\|_F^2/\|\Delta(\sigma)\|_F^2$.

\begin{wrapfigure}{r}{0.55\linewidth}
    \centering
    
    \includegraphics[width=\linewidth]{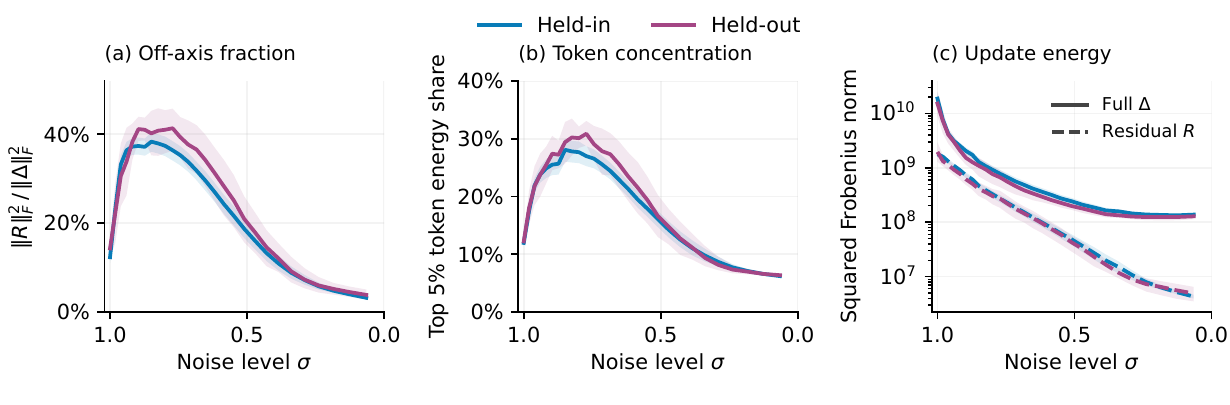}
        
    \caption{\textbf{Displacement field geometry varies along denoising.}
    Bands are pointwise 95\%
    case-bootstrap intervals. Noise decreases from left to right.}
    \label{fig:per_sigma_geometry}
    
\end{wrapfigure}
Fig.~\ref{fig:per_sigma_geometry} shows that both direction structure and the
distribution of energy across tokens change systematically during generation. For
held-in/held-out styles, $\rho$ starts at 12.3\%/14.3\%, reaches measured
maxima of 38.3\%/41.4\% at $\sigma=0.849/0.772$, and falls to 3.2\%/3.8\%
at $\sigma=0.065$.
The highest-energy 5\% of tokens carry a larger energy share in
the intermediate part of the trajectory, returning to about 6\% near its
end. Late updates are therefore more aligned across tokens. Suppression shows a similar pattern
(App.~\ref{app:per_sigma_geometry}).

These results show that the field adapts its update geometry along the generation trajectory beyond a uniform rescaling of a fixed displacement,
as $\Delta(\sigma)=a(\sigma)\Delta_{\mathrm{ref}}$ would leave both the off-axis
energy fraction and the top-token energy share unchanged for nonzero
$a(\sigma)$, unlike our observation.

\section{Conclusions and Limitations}
\label{sec:conclusion}

We introduced \ours, a concept-conditioned velocity field that steers image-token
activations at a single site in a frozen DiT. Its token-varying,
activation-dependent updates share learned parameters across concepts, enabling
fine-grained style descriptions and generalization to held-out concepts without
per-concept fitting. The integration horizon provides continuous control. Our experiments demonstrate strong style--content trade-offs and show the applicability to concept suppression.
The limitation is that \ours introduce computational overhead at inference time, which we discuss in App.~\ref{app:overhead}, and does not guarantee preservation of the original composition.
Intervention can change object positions, poses, and scene layout alongside the intended concept. Specific concept control remains an open challenge.

\section*{Ethics Statement}

This work uses existing datasets and pretrained text-to-image models and does not involve human-subject studies.
The proposed controller may inherit biases and harmful-generation capabilities from its pretrained backbones.
Although concept suppression can help reduce unwanted content, our NSFW-suppression results are limited to the evaluated settings and do not establish comprehensive safety against misuse.

\section*{Reproducibility Statement}

The model architecture and training objective are described in Section~\ref{sec:method}. Appendix~\ref{app:reproduction} provides backbone-specific implementation details, training hyperparameters, data construction procedures, and evaluation protocols.
The public repository provides training and evaluation code, experiment configurations, and data construction scripts for the style control and concept suppression experiments.

\section*{Data Availability Statement}

The public repository includes metadata of the training and test datasets for style control and concept suppression, specifying the source and target prompts and the concept description.
Scripts are provided to construct the training and evaluation targets from these resources.
A FLUX style-field checkpoint is included in the public code release. The frozen backbone weights are obtained separately under their original licenses.

\subsection*{AI use statement}
In this work, we used generative AI tools for generating synthetic datasets, assisting with translation, cleaning and reformatting dataset. We have not used generative AI tools for implementing methods, helping develop theoretical models or conceptual frameworks, formulating mathematical claims, providing critical ingredients for proving mathematical claims, proposing or refining hypotheses, designing or providing feedback on research methodology or experiments, supporting qualitative and thematic data analysis, or interpreting results, and assisting in the writing of proofs are not applicable to this work. Additionally, we used generative AI tools for creating and modifying scientific figures. We have checked all AI-assisted work by manually verifying LLM-generated code, data and figures for correctness by all authors. We take responsibility for the final content of this work, including text, claims or artifacts produced with the aid of generative AI.

\bibliography{references}

@article{chrabaszcz2026conditioned,
  title={Conditioned Activation Transport for {T2I} Safety Steering},
  author={Chrabaszcz, Maciej and Szymczyk, Aleksander and Dubi{\'n}ski, Jan and Trzci{\'n}ski, Tomasz and Boenisch, Franziska and Dziedzic, Adam},
  journal={arXiv preprint arXiv:2603.03163},
  year={2026},
  url={https://arxiv.org/abs/2603.03163}
}

@article{ferrando2025dsas,
  title={Dynamically Scaled Activation Steering},
  author={Ferrando, Alex and Suau, Xavier and Gonz{\`a}lez, Jordi and Rodriguez, Pau},
  journal={arXiv preprint arXiv:2512.03661},
  year={2025},
  url={https://arxiv.org/abs/2512.03661}
}

@article{zaleska2026attention,
  title={Attention, May I Have Your Decision? Localizing Generative Choices in Diffusion Models},
  author={Zaleska, Katarzyna and Popek, {\L}ukasz and Wysocza{\'n}ska, Monika and Deja, Kamil},
  journal={arXiv preprint arXiv:2604.06052},
  year={2026},
  url={https://arxiv.org/abs/2604.06052}
}

@inproceedings{rodriguez2025controlling,
  title={Controlling language and diffusion models by transporting activations},
  author={Rodriguez, Pau and Blaas, Arno and Klein, Michal and Zappella, Luca and Apostoloff, Nicholas and Suau, Xavier and others},
  booktitle={International Conference on Learning Representations},
  volume={2025},
  pages={89812--89855},
  year={2025}
}

@article{turner2024activation,
  title={Activation addition: Steering language models without optimization},
  author={Turner, Alexander Matt and Thiergart, Lisa and Leech, Gavin and Udell, David and Mini, Ulisse and MacDiarmid, Monte},
  year={2024}
}

@inproceedings{gaintseva2026casteer,
  title={CASteer: Cross-attention steering for controllable concept erasure},
  author={Gaintseva, Tatiana and Oncescu, Andreea-Maria and Ma, Chengcheng and Liu, Ziquan and Benning, Martin and Slabaugh, Gregory and Deng, Jiankang and Elezi, Ismail},
  booktitle={International Conference on Learning Representations},
  volume={2026},
  pages={40098--40110},
  year={2026}
}

@article{konovalova2026shift,
  title={SHIFT: Steering Hidden Intermediates in Flow Transformers},
  author={Konovalova, Nina and Kuznetsov, Andrey and Alanov, Aibek},
  journal={arXiv preprint arXiv:2604.09213},
  year={2026}
}

@article{li2026semantic,
  title={Semantic Steering for Controllable Generation: Tuning-Free Concept Erasure in Multimodal Diffusion Transformers},
  author={Li, Qiao and Fu, Xiaomeng and Zhao, Yuanshu and Wang, Qipeng and Dai, Jiao and Han, Jizhong},
  journal={arXiv preprint arXiv:2608.12829},
  year={2026}
}

@article{wu2026steeringdiffusion,
  title={SteeringDiffusion: A Bottlenecked Activation Control Interface for Diffusion Models},
  author={Wu, Fangzheng and Summa, Brian},
  journal={arXiv preprint arXiv:2605.01653},
  year={2026}
}

@inproceedings{lyu2024one,
  title={One-dimensional adapter to rule them all: Concepts, diffusion models and erasing applications},
  author={Lyu, Mengyao and Yang, Yuhong and Hong, Haiwen and Chen, Hui and Jin, Xuan and He, Yuan and Xue, Hui and Han, Jungong and Ding, Guiguang},
  booktitle={2024 IEEE/CVF Conference on Computer Vision and Pattern Recognition (CVPR)},
  pages={7559--7568},
  year={2024},
  organization={IEEE}
}

@article{smith2023continual,
  title={Continual diffusion: Continual customization of text-to-image diffusion with c-lora},
  author={Smith, James Seale and Hsu, Yen-Chang and Zhang, Lingyu and Hua, Ting and Kira, Zsolt and Shen, Yilin and Jin, Hongxia},
  journal={arXiv preprint arXiv:2304.06027},
  year={2023}
}

@inproceedings{tsai2024ring,
  title={Ring-a-bell! how reliable are concept removal methods for diffusion models?},
  author={Tsai, Yu-Lin and Hsu, Chia-Yi and Xie, Chulin and Lin, Chih-Hsun and Chen, Jia You and Li, Bo and Chen, Pin-Yu and Yu, Chia-Mu and Huang, Chun-Ying},
  booktitle={International Conference on Learning Representations},
  volume={2024},
  pages={41543--41554},
  year={2024}
}

@inproceedings{gandikota2024concept,
  title={Concept sliders: Lora adaptors for precise control in diffusion models},
  author={Gandikota, Rohit and Materzy{\'n}ska, Joanna and Zhou, Tingrui and Torralba, Antonio and Bau, David},
  booktitle={European Conference on Computer Vision},
  pages={172--188},
  year={2024},
  organization={Springer}
}

@inproceedings{chiu2026text,
  title={Text Slider: Efficient and Plug-and-Play Continuous Concept Control for Image/Video Synthesis via LoRA Adapters},
  author={Chiu, Pin-Yen and Fang, I and Chen, Jun-Cheng and others},
  booktitle={Proceedings of the IEEE/CVF Winter Conference on Applications of Computer Vision},
  pages={613--622},
  year={2026}
}

@inproceedings{nguyen2026activation,
  title={Activation steering with a feedback controller},
  author={Nguyen, Dung Viet and Pham, Yen and Vu, Hieu and Zhang, Lei and Nguyen, Tan},
  booktitle={International Conference on Learning Representations},
  volume={2026},
  pages={154471--154505},
  year={2026}
}

@inproceedings{ren2025six,
  title={Six-cd: Benchmarking concept removals for text-to-image diffusion models},
  author={Ren, Jie and Chen, Kangrui and Cui, Yingqian and Zeng, Shenglai and Liu, Hui and Xing, Yue and Tang, Jiliang and Lyu, Lingjuan},
  booktitle={2025 IEEE/CVF Conference on Computer Vision and Pattern Recognition (CVPR)},
  pages={28769--28778},
  year={2025},
  organization={IEEE}
}

@inproceedings{schramowski2023safe,
  title={Safe latent diffusion: Mitigating inappropriate degeneration in diffusion models},
  author={Schramowski, Patrick and Brack, Manuel and Deiseroth, Bj{\"o}rn and Kersting, Kristian},
  booktitle={2023 IEEE/CVF Conference on Computer Vision and Pattern Recognition (CVPR)},
  pages={22522--22531},
  year={2023},
  organization={IEEE}
}

@article{gao2026megastyle,
  title={Megastyle: Constructing diverse and scalable style dataset via consistent text-to-image style mapping},
  author={Gao, Junyao and Liu, Sibo and Li, Jiaxing and Sun, Yanan and Tu, Yuanpeng and Shen, Fei and Zhang, Weidong and Zhao, Cairong and Zhang, Jun},
  journal={arXiv preprint arXiv:2604.08364},
  year={2026}
}

@article{cai2025z,
  title={Z-image: An efficient image generation foundation model with single-stream diffusion transformer},
  author={Cai, Huanqia and Cao, Sihan and Du, Ruoyi and Gao, Peng and Hao, Aiming and Hoi, Steven and Hou, Zhaohui and Huang, Shijie and Jiang, Dengyang and Jiang, Yuming and others},
  journal={arXiv preprint arXiv:2511.22699},
  year={2025}
}

@article{hui2024hq,
  title={Hq-edit: A high-quality dataset for instruction-based image editing},
  author={Hui, Mude and Yang, Siwei and Zhao, Bingchen and Shi, Yichun and Wang, Heng and Wang, Peng and Zhou, Yuyin and Xie, Cihang},
  journal={arXiv preprint arXiv:2404.09990},
  year={2024}
}

@inproceedings{lin2014microsoft,
  title={Microsoft coco: Common objects in context},
  author={Lin, Tsung-Yi and Maire, Michael and Belongie, Serge and Hays, James and Perona, Pietro and Ramanan, Deva and Doll{\'a}r, Piotr and Zitnick, C Lawrence},
  booktitle={European conference on computer vision},
  pages={740--755},
  year={2014},
  organization={Springer}
}

@inproceedings{zhang2024generate,
  title={To generate or not? safety-driven unlearned diffusion models are still easy to generate unsafe images... for now},
  author={Zhang, Yimeng and Jia, Jinghan and Chen, Xin and Chen, Aochuan and Zhang, Yihua and Liu, Jiancheng and Ding, Ke and Liu, Sijia},
  booktitle={European Conference on Computer Vision},
  pages={385--403},
  year={2024},
  organization={Springer}
}

@inproceedings{baumann2025continuous,
  title={Continuous, subject-specific attribute control in t2i models by identifying semantic directions},
  author={Baumann, Stefan Andreas and Krause, Felix and Neumayr, Michael and Stracke, Nick and Sevi, Melvin and Hu, Vincent Tao and Ommer, Bj{\"o}rn},
  booktitle={2025 IEEE/CVF Conference on Computer Vision and Pattern Recognition (CVPR)},
  pages={13231--13241},
  year={2025},
  organization={IEEE}
}

@inproceedings{brooks2023instructpix2pix,
  title={Instructpix2pix: Learning to follow image editing instructions},
  author={Brooks, Tim and Holynski, Aleksander and Efros, Alexei A},
  booktitle={2023 IEEE/CVF Conference on Computer Vision and Pattern Recognition (CVPR)},
  pages={18392--18402},
  year={2023},
  organization={IEEE}
}

@inproceedings{ke2021musiq,
  title={Musiq: Multi-scale image quality transformer},
  author={Ke, Junjie and Wang, Qifei and Wang, Yilin and Milanfar, Peyman and Yang, Feng},
  booktitle={2021 IEEE/CVF International Conference on Computer Vision (ICCV)},
  pages={5128--5137},
  year={2021},
  organization={IEEE}
}

@article{oquab2023dinov2,
  title={Dinov2: Learning robust visual features without supervision},
  author={Oquab, Maxime and Darcet, Timoth{\'e}e and Moutakanni, Th{\'e}o and Vo, Huy and Szafraniec, Marc and Khalidov, Vasil and Fernandez, Pierre and Haziza, Daniel and Massa, Francisco and El-Nouby, Alaaeldin and others},
  journal={arXiv preprint arXiv:2304.07193},
  year={2023}
}

@inproceedings{zhang2018unreasonable,
  title={The unreasonable effectiveness of deep features as a perceptual metric},
  author={Zhang, Richard and Isola, Phillip and Efros, Alexei A and Shechtman, Eli and Wang, Oliver},
  booktitle={2018 IEEE/CVF conference on computer vision and pattern recognition},
  pages={586--595},
  year={2018},
  organization={IEEE}
}

@inproceedings{gandikota2024unified,
  title={Unified concept editing in diffusion models},
  author={Gandikota, Rohit and Orgad, Hadas and Belinkov, Yonatan and Materzy{\'n}ska, Joanna and Bau, David},
  booktitle={2024 IEEE/CVF Winter Conference on Applications of Computer Vision (WACV)},
  pages={5099--5108},
  year={2024},
  organization={IEEE}
}

@inproceedings{kumari2023ablating,
  title={Ablating concepts in text-to-image diffusion models},
  author={Kumari, Nupur and Zhang, Bingliang and Wang, Sheng-Yu and Shechtman, Eli and Zhang, Richard and Zhu, Jun-Yan},
  booktitle={2023 IEEE/CVF International Conference on Computer Vision (ICCV)},
  pages={22634--22645},
  year={2023},
  organization={IEEE}
}

@inproceedings{lu2024mace,
  title={Mace: Mass concept erasure in diffusion models},
  author={Lu, Shilin and Wang, Zilan and Li, Leyang and Liu, Yanzhu and Kong, Adams Wai-Kin},
  booktitle={2024 IEEE/CVF Conference on Computer Vision and Pattern Recognition (CVPR)},
  pages={6430--6440},
  year={2024},
  organization={IEEE}
}

@inproceedings{gandikota2023erasing,
  title={Erasing concepts from diffusion models},
  author={Gandikota, Rohit and Materzynska, Joanna and Fiotto-Kaufman, Jaden and Bau, David},
  booktitle={Proceedings of the IEEE/CVF international conference on computer vision},
  pages={2426--2436},
  year={2023}
}

@inproceedings{gao2025eraseanything,
  title={Eraseanything: Enabling concept erasure in rectified flow transformers},
  author={Gao, Daiheng and Lu, Shilin and Zhou, Wenbo and Chu, Jiaming and Zhang, Jie and Jia, Mengxi and Zhang, Bang and Fan, Zhaoxin and Zhang, Weiming},
  booktitle={Forty-second International Conference on Machine Learning},
  year={2025}
}

@inproceedings{peebles2023scalable,
  title={Scalable diffusion models with transformers},
  author={Peebles, William and Xie, Saining},
  booktitle={2023 IEEE/CVF International Conference on Computer Vision (ICCV)},
  pages={4172--4182},
  year={2023},
  organization={IEEE}
}

@inproceedings{radford2021learning,
  title={Learning transferable visual models from natural language supervision},
  author={Radford, Alec and Kim, Jong Wook and Hallacy, Chris and Ramesh, Aditya and Goh, Gabriel and Agarwal, Sandhini and Sastry, Girish and Askell, Amanda and Mishkin, Pamela and Clark, Jack and others},
  booktitle={International conference on machine learning},
  pages={8748--8763},
  year={2021},
  organization={PmLR}
}

@article{raffel2020exploring,
  title={Exploring the limits of transfer learning with a unified text-to-text transformer},
  author={Raffel, Colin and Shazeer, Noam and Roberts, Adam and Lee, Katherine and Narang, Sharan and Matena, Michael and Zhou, Yanqi and Li, Wei and Liu, Peter J},
  journal={Journal of machine learning research},
  volume={21},
  number={140},
  pages={1--67},
  year={2020}
}

@article{yang2025qwen3,
  title={Qwen3 technical report},
  author={Yang, An and Li, Anfeng and Yang, Baosong and Zhang, Beichen and Hui, Binyuan and Zheng, Bo and Yu, Bowen and Gao, Chang and Huang, Chengen and Lv, Chenxu and others},
  journal={arXiv preprint arXiv:2505.09388},
  year={2025}
}

@software{flux1dev2024,
  title        = {FLUX.1-dev},
  author       = {{Black Forest Labs}},
  year         = {2024},
  url          = {https://huggingface.co/black-forest-labs/FLUX.1-dev},
}

@software{nudnet,
  author = {Praneeth Bedapudi},
  title  = {NudeNet: Neural Nets for Nudity Classification, Detection and Selective Censoring},
  url    = {https://github.com/notAI-tech/NudeNet},
  year   = {2019}
}

@article{ho2022classifier,
  title={Classifier-free diffusion guidance},
  author={Ho, Jonathan and Salimans, Tim},
  journal={arXiv preprint arXiv:2207.12598},
  year={2022}
}

@article{hu2021lora,
  title={Lora: Low-rank adaptation of large language models},
  author={Hu, Edward J and Shen, Yelong and Wallis, Phillip and Allen-Zhu, Zeyuan and Li, Yuanzhi and Wang, Shean and Wang, Lu and Chen, Weizhu},
  journal={arXiv preprint arXiv:2106.09685},
  year={2021}
}

@inproceedings{liu2024grounding,
  title={Grounding dino: Marrying dino with grounded pre-training for open-set object detection},
  author={Liu, Shilong and Zeng, Zhaoyang and Ren, Tianhe and Li, Feng and Zhang, Hao and Yang, Jie and Jiang, Qing and Li, Chunyuan and Yang, Jianwei and Su, Hang and others},
  booktitle={European conference on computer vision},
  pages={38--55},
  year={2024},
  organization={Springer}
}

@inproceedings{kirillov2023segment,
  title={Segment anything},
  author={Kirillov, Alexander and Mintun, Eric and Ravi, Nikhila and Mao, Hanzi and Rolland, Chloe and Gustafson, Laura and Xiao, Tete and Whitehead, Spencer and Berg, Alexander C and Lo, Wan-Yen and others},
  booktitle={2023 IEEE/CVF international conference on computer vision (ICCV)},
  pages={3992--4003},
  year={2023},
  organization={IEEE}
}

@inproceedings{zhai2023sigmoid,
  title={Sigmoid loss for language image pre-training},
  author={Zhai, Xiaohua and Mustafa, Basil and Kolesnikov, Alexander and Beyer, Lucas},
  booktitle={2023 IEEE/CVF International Conference on Computer Vision (ICCV)},
  pages={11941--11952},
  year={2023},
  organization={IEEE}
}

@inproceedings{rombach2022high,
  title={High-resolution image synthesis with latent diffusion models},
  author={Rombach, Robin and Blattmann, Andreas and Lorenz, Dominik and Esser, Patrick and Ommer, Bj{\"o}rn},
  booktitle={2022 IEEE/CVF conference on computer vision and pattern recognition (CVPR)},
  pages={10674--10685},
  year={2022},
  organization={ieee}
}

@article{dalva2024fluxspace,
  title={Fluxspace: Disentangled semantic editing in rectified flow transformers},
  author={Dalva, Yusuf and Venkatesh, Kavana and Yanardag, Pinar},
  journal={arXiv preprint arXiv:2412.09611},
  year={2024}
}

@inproceedings{yang2026splitflux,
  title={SplitFlux: Learning to Decouple Content and Style from a Single Image},
  author={Yang, Yitong and Wang, Yinglin and Wang, Changshuo and Zhang, Yongjun and Chen, Ziyang and He, Shuting},
  booktitle={Proceedings of the IEEE/CVF Conference on Computer Vision and Pattern Recognition},
  pages={572--582},
  year={2026}
}

@inproceedings{rimsky2024steering,
  title={Steering llama 2 via contrastive activation addition},
  author={Rimsky, Nina and Gabrieli, Nick and Schulz, Julian and Tong, Meg and Hubinger, Evan and Turner, Alexander},
  booktitle={Proceedings of the 62nd Annual Meeting of the Association for Computational Linguistics (Volume 1: Long Papers)},
  pages={15504--15522},
  year={2024}
}

@article{chen2024expanding,
  title={Expanding performance boundaries of open-source multimodal models with model, data, and test-time scaling},
  author={Chen, Zhe and Wang, Weiyun and Cao, Yue and Liu, Yangzhou and Gao, Zhangwei and Cui, Erfei and Zhu, Jinguo and Ye, Shenglong and Tian, Hao and Liu, Zhaoyang and others},
  journal={arXiv preprint arXiv:2412.05271},
  year={2024}
}

@inproceedings{moon2025holistic,
  title={Holistic unlearning benchmark: A multi-faceted evaluation for text-to-image diffusion model unlearning},
  author={Moon, Saemi and Lee, Minjong and Park, Sangdon and Kim, Dongwoo},
  booktitle={2025 IEEE/CVF International Conference on Computer Vision (ICCV)},
  pages={16356--16366},
  year={2025},
  organization={IEEE}
}

@article{jin2026beyond,
  title={Beyond Steering Vector: Flow-based Activation Steering for Inference-Time Intervention},
  author={Jin, Zehao and Deng, Ruixuan and Wang, Junran and Shen, Xinjie and Zhang, Chao},
  journal={arXiv preprint arXiv:2605.05892},
  year={2026}
}
\bibliographystyle{acflow_preprint}

\appendix

\clearpage

\section{Computational Overhead}
\label{app:overhead}

\ours adds a learned module to a frozen generator, so we measure what this
costs at inference.

\textbf{Parameter count and computational overhead.}
The field has 0.25B parameters, 2.1\% of the 11.9B-parameter FLUX.1-dev transformer, and one velocity evaluation costs 0.30T FLOPs, 1.7\% of one DiT forward pass, since it is a single transformer block acting on the image tokens only. The field is evaluated $N$ times at one intervention site in every denoising step. All other computation is that of the unmodified DiT. FLOPs therefore grow linearly in $N$.

\textbf{Performance measurements.}
Table~\ref{tab:overhead} reports parameters, FLOPs, and end-to-end latency for $N\in\{1,3,5\}$ and batch size ${1,8}$ under the evaluation protocol of Sec.~\ref{sec:megastyle}, measured on one A100-80GB GPU.
The concept description is encoded once per generation (29\,ms) and reused across all denoising and integration steps, so it is excluded from the per-step cost.
All quality results in this paper run the field and the generator in \texttt{bf16}.

\begin{table}[h]
\caption{\textbf{Compute overhead of \ours relative to the base DiT.}
FLUX.1-dev at $512\times512$, 28 denoising steps, guidance 3.5,
\ours at single-stream block 16 with MegaStyle prompts and style
descriptions, one A100-80GB. FLOPs cover the full denoising loop.
Latency is the mean end-to-end time of one pipeline call from
precomputed text embeddings, including 28 denoising steps and VAE
decoding, over 10 calls after 2 warm-up calls
(standard deviation $\le 7$\,ms in every cell).
\texttt{bf16} is the precision used for all reported results.}
\label{tab:overhead}

\centering
\small
\setlength{\tabcolsep}{4pt}
\renewcommand{\arraystretch}{1.12}

\resizebox{\textwidth}{!}{\begin{tabular}{@{}clcccccc@{}}
\toprule
&
&
&
\multicolumn{3}{c}{\textsc{FLOPs}}
&
\multicolumn{2}{c}{\textsc{Latency (s)}} \\
\cmidrule(lr){4-6}
\cmidrule(l){7-8}

\textsc{Batch}
&
\textsc{Model}
&
\textsc{Params}
&
\textsc{Per step (T)}
&
\textsc{Total (P)}
&
$\Delta$
&
\textsc{Per call}
&
$\Delta$
\\

\midrule

\multirow{4}{*}{1}
& Base DiT
& 11.90B
& 17.69
& 0.495
& --
& 3.58
& --
\\

& \ours, $N{=}1$
& +0.25B
& 17.98
& 0.504
& +1.7\%
& 3.66
& +2.2\%
\\

& \cellcolor{oursbg}\ours, $N{=}3$
& \cellcolor{oursbg}+0.25B
& \cellcolor{oursbg}18.58
& \cellcolor{oursbg}0.520
& \cellcolor{oursbg}+5.0\%
& \cellcolor{oursbg}3.80
& \cellcolor{oursbg}+6.1\%
\\

& \ours, $N{=}5$
& +0.25B
& 19.17
& 0.537
& +8.4\%
& 3.93
& +9.8\%
\\

\midrule

\multirow{4}{*}{8}
& Base DiT
& 11.90B
& 141.5
& 3.96
& --
& 23.82
& --
\\

& \ours, $N{=}1$
& +0.25B
& 143.8
& 4.03
& +1.7\%
& 24.27
& +1.9\%
\\

& \cellcolor{oursbg}\ours, $N{=}3$
& \cellcolor{oursbg}+0.25B
& \cellcolor{oursbg}148.6
& \cellcolor{oursbg}4.16
& \cellcolor{oursbg}+5.0\%
& \cellcolor{oursbg}25.11
& \cellcolor{oursbg}+5.4\%
\\

& \ours, $N{=}5$
& +0.25B
& 153.4
& 4.30
& +8.4\%
& 25.96
& +9.0\%
\\

\bottomrule
\end{tabular}}

\end{table}

\section{Portability Across Backbones}
\label{app:z}

We instantiate the same activation-flow interface on Z-Image~\citep{cai2025z}
with separately trained, backbone-matched fields. The style field intervenes
after layer 19 and the suppression field after layer 25 (zero-based indices);
both modify image tokens on the positive CFG~\citep{ho2022classifier} branch. The backbone remains
frozen. We train the style control field on the same MegaStyle train split as with FLUX and the concept suppression field on the removal split from HQ-Edit~\citep{hui2024hq}. Implementation and actual training budgets are in App.~\ref{app:reproduction}.

\begin{figure}[htbp]
\centering
\includegraphics[width=0.9\linewidth]{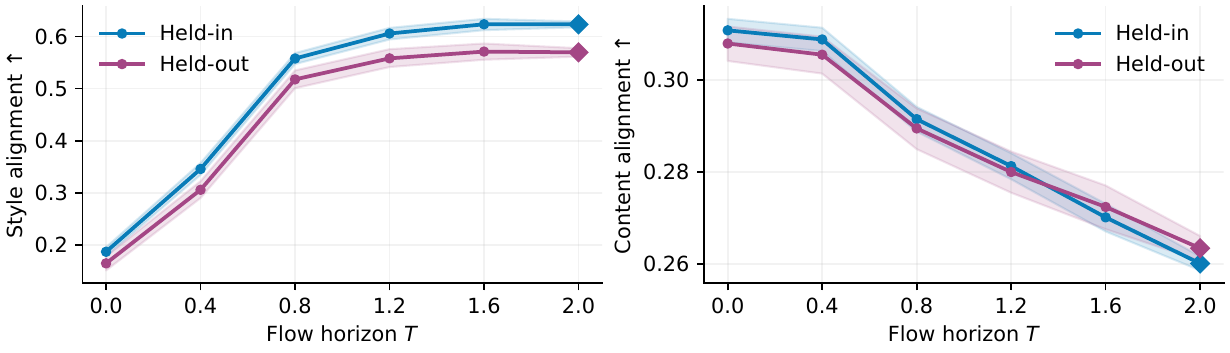}
\caption{\textbf{Style control on Z-Image}
evaluated on held-in and held-out style families.
Bands are pointwise 95\% intervals from 2,000
within-family stratified-bootstrap resamples.}
\label{fig:zimage_quantitative}
\end{figure}

\textbf{Style control.}
We evaluate at $T\in\{0,0.4,0.8,1.2,1.6,2.0\}$,
using the same fine-grained style descriptions, content prompts and scoring protocol as the FLUX experiment:
the full 2,048/768 samples at $T=2$ and quarter 512 subsets at others.
Figure~\ref{fig:zimage_quantitative} shows a consistent response to the
flow horizon on both splits. On the fixed subsets, increasing $T$
from 0 to 0.8 raises style alignment from 0.1868 to 0.5579 on held-in
styles and from 0.1645 to 0.5178 on held-out styles, while content alignment
changes from 0.3108 to 0.2915 and from 0.3079 to 0.2895, respectively.
Style alignment begins to saturate at larger horizons while content
alignment continues to decline. These results demonstrate controllable
style control and generalization to unseen style families on a second
backbone with a separately trained field.

\begin{figure}[h]
\centering
\includegraphics[width=\linewidth]{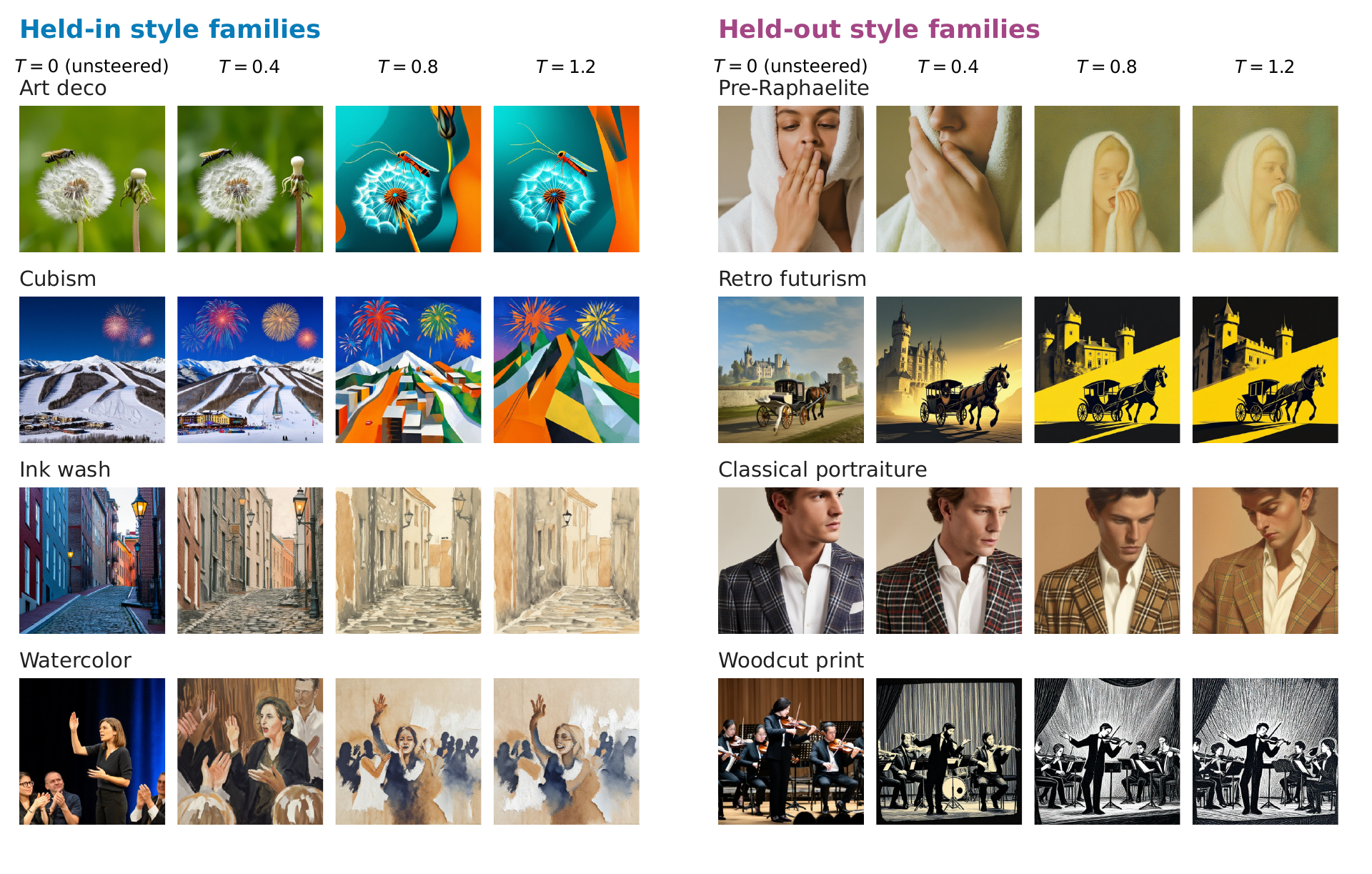}
\caption{\textbf{Qualitative style control results on Z-Image.}
Four held-in and four held-out style families, using the same style field. Each row varies the flow horizon $T\in\{0,0.4,0.8,1.2\}$ with a fixed prompt,
style description, and initial noise. Family names label the rows;
the field receives the corresponding fine-grained descriptions.}
\label{fig:zimage_style_selected}
\label{fig:zimage_portability}
\end{figure}

Figure~\ref{fig:zimage_style_selected} presents four held-in and four
held-out style families. Each row keeps its content prompt, fine-grained style description,
and seed fixed while increasing the flow horizon.

\begin{figure}[htbp]
\centering
\includegraphics[width=\linewidth]{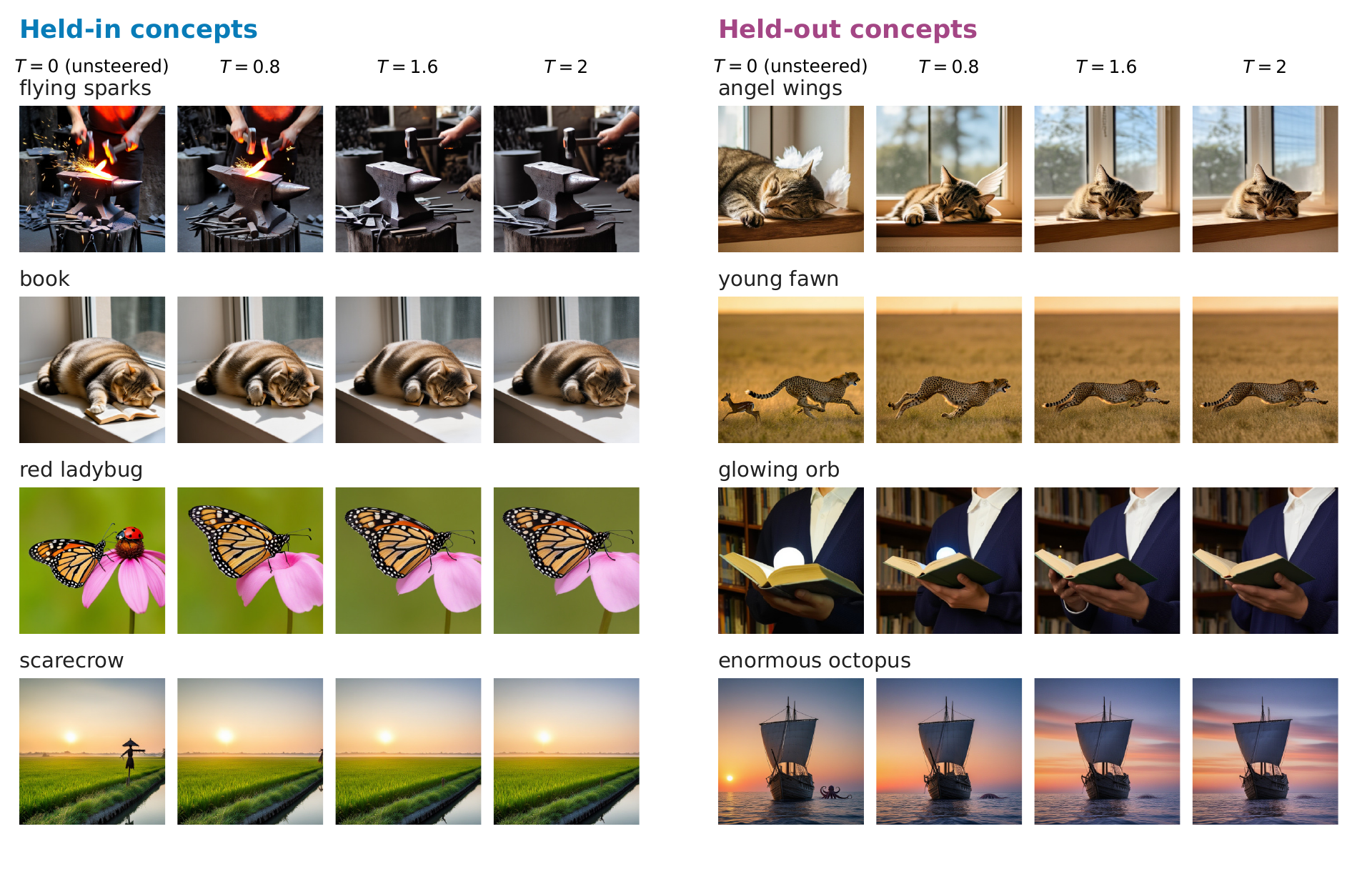}
\caption{\textbf{Qualitative concept suppression results on Z-Image.}
Four held-in and four held-out targets, using the same suppression field. Concept names label the removal target; prompts and initial
noise are fixed within each row.}
\label{fig:zimage_suppression_selected}
\end{figure}

\textbf{Concept suppression.}
Figure~\ref{fig:zimage_suppression_selected} presents four held-in and four
held-out targets. The selected examples show the target
present in the unintervened image and absent at larger horizons, with the
remaining scene still recognizable. We show more results in App.~\ref{app:z_full_scans}.

\section{Complete Results on Concept Suppression}
\label{app:ip2p_quantitative}

This section reports the complete evaluation of the suppression field of Sec.~\ref{sec:concept_negation}. It covers the protocol, the full
results (Table~\ref{tab:ip2p_suppression}), what the concept-free reference row does and
does not mean, why these prompts are hard to suppress, and the controls and limits of
the VLM judge. The evaluation contains 150 held-out concepts given as text at inference, whereas per-concept erasure methods fit one edit per concept.

\paragraph{Protocol.}
We use the same single field as in Sec.~\ref{sec:concept_negation}
and evaluate every row of both test splits: the held-in split has 149 prompts over 91
concepts that occur in training, with no prompt shared with training, and the held-out
split has 173 prompts over 150 concepts that never occur in training. The frozen
generator always receives the original prompt, which names the unwanted concept, and the
field receives \texttt{erase <concept>}. For each row we generate four images at
$T\in\{0,1,2,3\}$ with the standard settings of Sec.~\ref{sec:experiments} from the same initial noise.

\paragraph{Metrics.}
\emph{Presence} is the fraction of images for which
InternVL2.5-8B-MPO~\citep{chen2024expanding} answers ``yes'' to the question ``Is the
following concept clearly visible in the image: \texttt{<concept>}? Answer with exactly
one word: yes or no.'' \emph{Suppr.} conditions on the prompts whose
unsteered image is judged to show the concept, 142 prompts in each split, and reports the
fraction whose steered image is not. CLIP is the ViT-B/32~\citep{radford2021learning} cosine similarity
between the image and the prompt with the concept removed, which describes what should
remain, and MUSIQ~\citep{ke2021musiq} is a no-reference quality score that
detects image collapse.

\begin{figure}
    \centering
    \includegraphics[width=\linewidth]{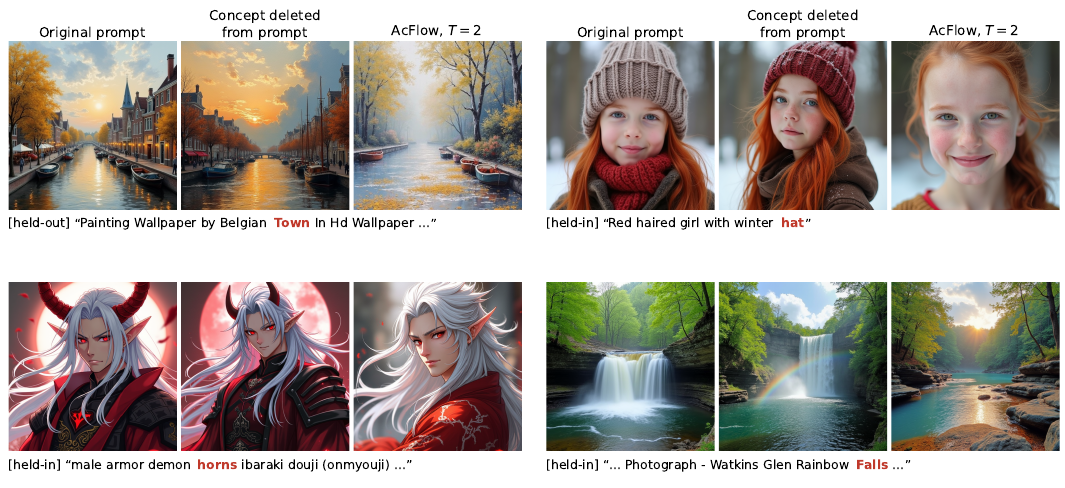}
    \caption{\textbf{Suppression where deleting the concept from the prompt fails.} Selected test cases in which the concept (red) remains visible after it is deleted from the prompt, because the remaining text still implies it, while \ours at $T{=}2$ removes it. The three images of a case share the initial noise.}
    \label{fig:prompt_editing_failure}
\end{figure}

\paragraph{Where directly prompt editing fails.}
While deleting a word stops asking for the concept, it does not ask for its absence as the context may still imply the concept.
To show that the test prompts are demanding in themselves, we generate reference images with the vanilla FLUX.1-dev from the original prompt with the concept phrase deleted, i.e., the concept-free caption that InstructPix2Pix pairs with each prompt, as is shown in Fig.~\ref{fig:prompt_editing_failure}. Most of these captions are plain deletions; a few replace the phrase
instead (e.g., ``lifeboat station'' becomes ``empty field'').
It is not a suppression method as it needs a rewritten prompt for every image, while \ours consumes only a text description of the concept to be removed.
The concept is still judged visible in 28.9\% of held-in and 15.6\% of held-out reference images, and among prompts whose unsteered image shows the concept, deleting it from the prompt removes it in only 69.7\%
and 83.8\% of cases.
For \ours these cases are harder than for the reference, since the generator is
still explicitly prompted with the concept and the field must override both the word and
the context that implies it.

\begin{table}[h]
\centering
\caption{\textbf{Concept suppression on IP2P.} Presence and Suppr. are percentages from the
InternVL2.5-8B-MPO judge; CLIP is the similarity to the concept-free prompt and MUSIQ is
no-reference image quality. All \ours rows and FLUX.1-dev share the initial noise of each
prompt. \emph{Reference} is unsteered FLUX.1-dev generating from the concept-free prompt. Held-out concepts are disjoint
from all training concepts.}
\label{tab:ip2p_suppression}
\small
\setlength{\tabcolsep}{4.5pt}
\begin{adjustbox}{max width=\linewidth}
\begin{tabular}{l cccc cccc}
\toprule
 & \multicolumn{4}{c}{Held-in} & \multicolumn{4}{c}{Held-out} \\
\cmidrule(lr){2-5}\cmidrule(lr){6-9}
 & Presence\,$\downarrow$ & Suppr.\,$\uparrow$ & CLIP\,$\uparrow$ & MUSIQ\,$\uparrow$
 & Presence\,$\downarrow$ & Suppr.\,$\uparrow$ & CLIP\,$\uparrow$ & MUSIQ\,$\uparrow$ \\
\midrule
FLUX.1-dev ($T{=}0$) & 95.3 & --   & 0.286 & 72.2 & 82.1 & --   & 0.283 & 72.6 \\
\ours, $T{=}1$      & 68.5 & 29.6 & 0.287 & 70.9 & 59.0 & 31.0 & 0.283 & 72.1 \\
\ours, $T{=}2$      & 41.6 & 57.0 & 0.273 & 69.7 & 40.5 & 54.2 & 0.267 & 70.4 \\
\ours, $T{=}3$      & 34.9 & 64.1 & 0.254 & 64.8 & 37.6 & 58.5 & 0.251 & 64.6 \\
\midrule
Reference & 28.9 & 69.7 & 0.302 & 71.3 & 15.6 & 83.8 & 0.298 & 71.5 \\
\bottomrule
\end{tabular}
\end{adjustbox}
\end{table}

\paragraph{Results.}
Presence decreases with the flow horizon on both splits, from 95.3\% to 34.9\% on held-in
and from 82.1\% to 37.6\% on held-out prompts. At $T{=}2$ the field suppresses the concept
in 57.0\% of held-in cases (95\% row-level bootstrap interval 48.6--65.5) and 54.2\% of
held-out cases (46.5--62.0); at $T{=}3$ the rates are 64.1\% (56.3--71.8) and 58.5\%
(50.7--66.2). The held-in and held-out rates lie inside each other's intervals at every
horizon, so we find no measurable loss when the concept is new to the field. The
preservation columns motivate $T{=}2$ as the operating point. Relative to the paired
unsteered image, CLIP changes by $-0.013$/$-0.016$ and MUSIQ by $-2.4$/$-2.2$ at $T{=}2$;
at $T{=}1$ CLIP is unchanged and MUSIQ drops by at most $1.3$. At $T{=}3$ suppression
improves by only 4--7 points while CLIP falls by $0.033$/$0.032$ and MUSIQ by
$7.4$/$8.0$. These images remain coherent, but they are softer and drift further from the
described scene, which matches the composition changes visible in
Fig.~\ref{fig:ip2p_strength}.

\begin{table}[h]
\centering
\caption{\textbf{Examples of prompts on which deleting the concept from the text does not
remove it.} Struck-through text is removed in the concept-free prompt. The last columns
give the judge's verdict on whether the concept is visible (\cmark) or not (\xmark) for
the reference and for \ours, which still receives the full original prompt.}
\label{tab:ip2p_hard_prompts}
\small
\setlength{\tabcolsep}{4pt}
\begin{tabular}{l p{0.47\linewidth} c ccc}
\toprule
Concept & Original prompt & Ref. & $T{=}1$ & $T{=}2$ & $T{=}3$ \\
\midrule
\multicolumn{6}{l}{\emph{(i) Proper name carries the concept}} \\
Mountains & wide angle view of Torngat \sout{Mountains} & \cmark & \cmark & \cmark & \cmark \\
River & Photograph - Kicking Horse \sout{River} by Pierre Leclerc Photography & \cmark & \cmark & \cmark & \xmark \\
Bridge & ``Bay \sout{Bridge} at Dawn'' Location: Embarcadero Plaza, San Francisco & \cmark & \cmark & \cmark & \xmark \\
Beach & Hoyvika \sout{Beach} on And{\o}ya, Vesteralen & \cmark & \xmark & \xmark & \xmark \\
\midrule
\multicolumn{6}{l}{\emph{(ii) Scene implies the concept}} \\
snow & Picture winter, \sout{snow,} trees, landscape, nature, fog, street, home & \cmark & \cmark & \xmark & \cmark \\
mist & Buttermere morning \sout{mist} & \cmark & \cmark & \xmark & \cmark \\
snow & Moraine Lake after a night of rain \sout{and snow} & \cmark & \cmark & \cmark & \cmark \\
\midrule
\multicolumn{6}{l}{\emph{(iii) Mined concept is not a removable object}} \\
Watercolor & WARSAW POLAND Original 7.5x11.5 Ink \sout{and Watercolor} Painting & \cmark & \cmark & \xmark & \xmark \\
Nature & Handmade Oil Paintings \sout{Of Nature} For Home Office Decoration & \cmark & \cmark & \cmark & \xmark \\
\bottomrule
\end{tabular}
\end{table}

\paragraph{Suppression where prompt editing fails.}
Table~\ref{tab:ip2p_reference_split} splits the conditioned prompts by whether the
reference removes the concept. Where prompt editing succeeds, \ours reaches 63.6\%/57.1\%
at $T{=}2$. Where prompt editing
fails, \ours is weaker at $T{=}2$ (41.9\%/39.1\%) but still suppresses 60.5\%/73.9\% at
$T{=}3$. The field thus can remove concepts that remain when the word is deleted
from the prompt. Table~\ref{tab:ip2p_hard_prompts} also
shows that individual verdicts are not always monotone in $T$; the monotone trend holds
for the aggregate rates.

\begin{table}[h]
\centering
\caption{\textbf{Suppression rate of \ours (\%) split by the outcome of prompt editing.}
Rows partition the prompts whose unsteered image shows the concept according to whether
the concept-free reference still shows it.}
\label{tab:ip2p_reference_split}
\small
\begin{tabular}{l cccc cccc}
\toprule
 & \multicolumn{4}{c}{Held-in} & \multicolumn{4}{c}{Held-out} \\
\cmidrule(lr){2-5}\cmidrule(lr){6-9}
Reference outcome & $n$ & $T{=}1$ & $T{=}2$ & $T{=}3$ & $n$ & $T{=}1$ & $T{=}2$ & $T{=}3$ \\
\midrule
Concept removed & 99 & 34.3 & 63.6 & 65.7 & 119 & 33.6 & 57.1 & 55.5 \\
Concept remains & 43 & 18.6 & 41.9 & 60.5 & 23 & 17.4 & 39.1 & 73.9 \\
\midrule
All & 142 & 29.6 & 57.0 & 64.1 & 142 & 31.0 & 54.2 & 58.5 \\
\bottomrule
\end{tabular}
\end{table}

\paragraph{Judge control and limitations.}
To check that the judge does not answer ``yes'' indiscriminately, we pair every unsteered
image with the concept of a different row of the same split. The judge answers ``yes''
in 10.7\% of held-in and 5.8\% of held-out pairs. This is an upper bound on its
false-positive rate, since frequent concepts such as sky or snow are genuinely present in
many other images. Two limits remain. The verdict is binary and the question asks for
\emph{clear} visibility, so partial attenuation, such as the smaller lighthouse or
narrower river in Fig.~\ref{fig:ip2p_strength}, counts as a failure; the table therefore
understates the graded response that the flow horizon provides. Finally, CLIP and MUSIQ show that suppression
is not free: \ours lowers the alignment with the remaining content slightly, and it does not guarantee that content unrelated to
the concept is left untouched.

\section{Intervention Site Diagnostic }
\label{app:intervention_details}

We use a token-swap oracle to test whether the image-token residual selected in
Section~\ref{sec:image_tokens} is causally sufficient for the target edit.
The diagnostic uses FLUX.1-dev at $512\times512$ resolution, 28 denoising
steps, embedded guidance 3.5, and single-stream block 16.  We sample eight
fixed rows from each of the 39 MegaStyle training families, for 312 prompts in
total.

\paragraph{Protocol.}
For each row, the source trajectory uses the content-only prompt and the target
trajectory appends its style description.  Both trajectories and all swap
conditions reuse an explicitly saved initial latent.  At every denoising step,
we record the target-prompt output of block 16 and replace either its text-token
slice, its image-token slice, or both slices in the source trajectory.  The
direct target-prompt generation defines the attainable style effect. We use the same style and content alignment metrics as in Sec.~\ref{sec:megastyle}. For a
condition $m$, we compute recovery from the dataset-level mean style score $q$,
\begin{equation}
    R_m =
    \frac{\bar q_m-\bar q_{\mathrm{plain}}}
         {\bar q_{\mathrm{target}}-\bar q_{\mathrm{plain}}}.
    \label{eq:oracle_recovery}
\end{equation}
Intervals use a two-level paired bootstrap that first resamples style families
and then prompts within each selected family (10,000 repetitions).

\begin{table}[h]
\centering\small
\caption{\textbf{Block-16 token-swap diagnostic on 312 MegaStyle prompts.}
Brackets give 95\% paired hierarchical-bootstrap intervals.
Recovery is relative to the direct target-prompt style gain.}
\label{tab:image_token_oracle}
\setlength{\tabcolsep}{4pt}
\begin{tabular}{@{}lccc@{}}
\toprule
Condition & Style score $\uparrow$ & Effect recovery & Content $\uparrow$ \\
\midrule
Plain prompt & 0.1714 [0.1543, 0.1876] & -- & 0.2982 \\
Direct target prompt & 0.5522 [0.5274, 0.5777] & 100\% (reference) & 0.2860 \\
Text tokens only & 0.3468 [0.3279, 0.3666] & 46.1\% [41.0, 51.3] & 0.2983 \\
Image tokens only & 0.5270 [0.5013, 0.5535] & 93.4\% [91.5, 95.1] & 0.2866 \\
Both token groups & 0.5512 [0.5266, 0.5769] & 99.7\% [99.4, 100.2] & 0.2861 \\
\bottomrule
\end{tabular}
\end{table}

Image-only replacement recovers 93.4\% of the target-prompt style gain,
compared with 46.1\% for text-only replacement. Their paired difference is
47.3 percentage points (95\% CI: 41.8--52.5). Replacing both groups recovers
99.7\%, 6.4 points above image-only. It supports the conclusion that the image-state path
carries most of the target effect at this block. These motivate the image-state intervention.

\section{Extended Flow Field Analysis}
\label{app:token_geometry_controls}

\subsection{Token-Direction and Energy Controls}
\label{app:per_token_geometry}

\newcommand{\supCases}{143 held-in and 146 held-out cases}
\newcommand{\supEnergy}{88.7\%/90.7\%}        \newcommand{\supResidualEnergy}{11\%/9\%}    \newcommand{\supFullRate}{53.8\%/54.1\%}      \newcommand{\supRankOne}{15.6\%/29.1\%}        \newcommand{\supResidual}{26.0\%/50.6\%}      \newcommand{\supResidualFull}{100.0\%/108.9\%} 

\paragraph{Protocol and aggregation.}
We evaluate the default style field at $T=2$ on 39 held-in cases, one per style family,
and on 36 held-out cases, six prompts for each of the six held-out families. The suppression field of Sec.~\ref{sec:concept_negation} is evaluated at $T=2$
on every prompt of the two IP2P evaluation sets whose unsteered image shows the concept,
which gives \supCases{}. Cases were randomly chosen from unsteered and full-field images only. The uncentered rank-one component in
Eq.~\ref{eq:token_rank1} allows a separate signed coefficient for each token;
it is more permissive than a constant vector broadcast to every token.
Each counterfactual recomputes the field and its axis on its own current
state at every denoising step. Norm-matched controls are rescaled separately
at each step to the full displacement norm at that same state.

\begin{figure}[h]
    \centering
    \includegraphics[width=\linewidth]{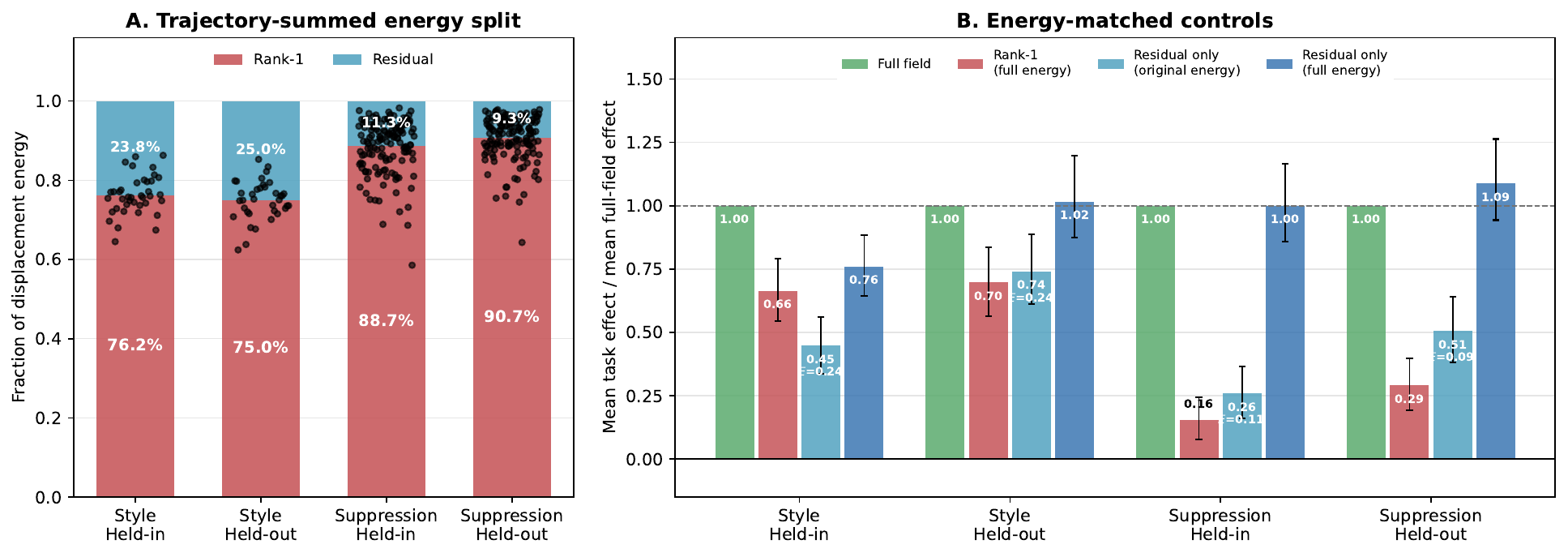}
    \caption{\textbf{Complete token-direction energy controls.}
    (A) Case-mean fractions of trajectory-summed displacement energy
    (Eq.~\ref{eq:token_energy_aggregation}).
    (B) Mean intervention effect divided by mean full-field effect, with
    95\% paired case-bootstrap intervals. Full-energy controls match norms
    separately at every denoising step; $E$ reports the original-norm
    residual's trajectory-summed applied/full energy ratio on its own trajectory.
    Style uses the reference-image MegaStyle cosine gain; suppression uses the removal of a
    concept that is visible in the unsteered image, as judged by InternVL2.5-8B-MPO.}
    \label{fig:token_energy_controls}
\end{figure}

Let $j$ index the $M$ cases in a split and $k$ index the 28 denoising steps. The energy fraction
in Fig.~\ref{fig:token_energy_controls}A is measured on full-field trajectories:
\begin{equation}
    \overline{e}_{Q_1}
    =\frac{1}{M}\sum_{j=1}^{M}
    \frac{\sum_{k=1}^{28}\|Q_{1,jk}\|_F^2}
         {\sum_{k=1}^{28}\|\Delta_{jk}\|_F^2}.
    \label{eq:token_energy_aggregation}
\end{equation}
Thus, steps are weighted by their full-field energy within each case,
while cases receive equal weight. This is neither a single-step statistic
nor an unweighted average of stepwise fractions.

\paragraph{Effect measures.}
For intervention $m$ and case $j$, $g_j^{(m)}$ is the effect relative to the unsteered
image with the same prompt and seed. For style it is the gain in MegaStyle
reference-image cosine. Suppression uses the judge of
App.~\ref{app:ip2p_quantitative}. Every case shows the concept in the unsteered image, and
$g_j^{(m)}\in\{0,1\}$ is one if the concept is no longer judged visible.
Figure~\ref{fig:token_energy_controls}B and Table~\ref{tab:token_ablation} report
$\overline{g}^{(m)}/\overline{g}^{(\mathrm{full})}$ within each split. The 95\% percentile intervals use 10,000 paired case-bootstrap
resamples for both tasks.

\begin{table}[t]
\centering\small
\caption{\textbf{Intervention ablation for style and suppression.} Mean effect divided by
mean full-field effect; brackets are 95\% paired case-bootstrap intervals. The style effect
is the MegaStyle style gain and the suppression effect is the VLM-judged removal of the
concept. The last rows give the rank-one share of the trajectory-summed displacement
energy and the energy applied by the learned-norm residual relative to the full update.
Rank-one directions are refitted on each live
trajectory; full-norm controls match the full update at every step.}
\label{tab:token_ablation}
\label{tab:style_token_ablation}
\begin{adjustbox}{max width=\linewidth}
\begin{tabular}{@{}lcccc@{}}
\toprule
 & \multicolumn{2}{c}{Style} & \multicolumn{2}{c}{Suppression} \\
\cmidrule(lr){2-3}\cmidrule(lr){4-5}
Update & Held-in & Held-out & Held-in & Held-out \\
\midrule
Rank one, full norm & 0.664 [0.545, 0.793] & 0.698 [0.565, 0.837] & 0.156 [0.079, 0.244] & 0.291 [0.192, 0.397] \\
Residual, learned norm & 0.448 [0.336, 0.561] & 0.741 [0.612, 0.886] & 0.260 [0.162, 0.365] & 0.506 [0.383, 0.640] \\
Residual, full norm & 0.759 [0.644, 0.884] & 1.016 [0.875, 1.197] & 1.000 [0.859, 1.164] & 1.089 [0.943, 1.263] \\
\midrule
Rank-one energy share & 76.2\% & 75.0\% & 88.7\% & 90.7\% \\
Residual energy applied & 24\% & 24\% & 11\% & 9\% \\
\bottomrule
\end{tabular}
\end{adjustbox}
\end{table}

\paragraph{Energy and intervention effect.}
Rank one accounts for 76.2\%/75.0\% of full-trajectory displacement energy on
held-in/held-out styles and for \supEnergy{} on held-in/held-out suppression concepts.
Matching its norm to the full update does not recover the full field: it retains
66.4\%/69.8\% of the mean style gain and \supRankOne{} of the suppression effect, where
the full field removes the concept in \supFullRate{} of the cases. What rank one lacks is
therefore direction, not magnitude.

The residual is not discarded noise. At its learned
norm it applies about a quarter of the update energy for style and \supResidualEnergy{}
for suppression, yet it retains 44.8\%/74.1\% of the style gain and \supResidual{} of the
suppression effect, and every interval excludes zero. Rescaled to the full norm it
recovers 75.9\%/101.6\% and \supResidualFull{}, i.e., the residual directions alone
reproduce the full-field suppression effect.  The rescaled residual does not degrade the images: its mean MUSIQ is 70.7/72.9 on held-in/held-out styles and 72.0/72.3 on suppression, against 71.2/72.9 and 70.9/70.8 for the full field. Captured displacement energy thus does not
predict effect.

Off-axis directions matter in both tasks and most for suppression, where
the shared axis holds about nine tenths of the energy but less than a third of the
effect. A plausible reason might be that suppression must change the tokens that depict the
object differently from the rest of the scene, which a single direction with per-token
coefficients represents poorly, whereas style changes most tokens in a more similar yet still different way, which we leave as a future direction.

\paragraph{Image quality and content.}
Table~\ref{tab:token_controls_quality} checks that no effect reported above comes from
damaged images. Mean MUSIQ stays within about two points of the unsteered images for
every control, and the residual rescaled to the full norm matches the full field in both
MUSIQ and CLIP similarity to the text describing the remaining content. It therefore
reproduces the full-field effect at the same cost as the full field and not by destroying
the image. The rank-one control has the highest MUSIQ and stays closest to the unsteered
images, so its weak effect is not caused by image degradation either.

\begin{table}[t]
\centering\small
\caption{\textbf{Image quality and content alignment of the controls}, on the cases of
Table~\ref{tab:token_ablation} (for suppression, all 149/173 prompts). MUSIQ is
no-reference image quality. CLIP is the ViT-B/32 cosine similarity to the content prompt
for style and to the concept-free prompt for suppression.}
\label{tab:token_controls_quality}
\begin{adjustbox}{max width=\linewidth}
\begin{tabular}{@{}lcccccccc@{}}
\toprule
 & \multicolumn{4}{c}{MUSIQ $\uparrow$} & \multicolumn{4}{c}{CLIP to remaining content $\uparrow$} \\
\cmidrule(lr){2-5}\cmidrule(lr){6-9}
 & \multicolumn{2}{c}{Style} & \multicolumn{2}{c}{Suppression} & \multicolumn{2}{c}{Style} & \multicolumn{2}{c}{Suppression} \\
\cmidrule(lr){2-3}\cmidrule(lr){4-5}\cmidrule(lr){6-7}\cmidrule(lr){8-9}
Update & Held-in & Held-out & Held-in & Held-out & Held-in & Held-out & Held-in & Held-out \\
\midrule
Unsteered & 71.9 & 73.8 & 72.5 & 72.9 & 0.299 & 0.295 & 0.288 & 0.284 \\
Full field & 71.2 & 72.9 & 70.9 & 70.8 & 0.284 & 0.275 & 0.279 & 0.268 \\
\midrule
Rank one, full norm & 73.0 & 74.6 & 72.1 & 71.9 & 0.287 & 0.291 & 0.284 & 0.281 \\
Residual, learned norm & 71.8 & 73.3 & 72.0 & 72.4 & 0.294 & 0.286 & 0.284 & 0.284 \\
Residual, full norm & 70.7 & 72.9 & 72.0 & 72.3 & 0.283 & 0.274 & 0.275 & 0.268 \\
\bottomrule
\end{tabular}
\end{adjustbox}
\end{table}

\paragraph{Axis robustness and scope.}
We also refit after normalizing every nonzero token update to unit norm and
after excluding the highest-norm 10\% of tokens. These controls test whether
large-displacement tokens determine the shared axis. Mean absolute axis
cosines are 0.973/0.972 for equal-token PCA and 0.981/0.979 after trimming
on held-in/held-out styles, respectively, averaging over each case's 28-step
trajectory first.
Our conclusion concerns the energy-optimal shared component of the learned
field, adapted per sample, step, and trajectory. It does not rule out an
arbitrary direction selected by a different downstream objective.

\subsection{Geometry Across Denoising Noise Levels}
\label{app:per_sigma_geometry}

The per-noise-level analysis uses only full-field trajectories of 39 held-in and six held-out style
cases, plus 16 held-in and eight held-out suppression cases, with one case
per concept or style family. All trajectories contain 28 recorded
interventions at a common strictly decreasing $\sigma$ grid from 1 to
0.065. Both fields use block 16, Euler-3, and $T=2$.

For each case and step, we compute $\rho=\|R\|_F^2/\|\Delta\|_F^2$
using the step's own uncentered rank-one decomposition. The token
concentration statistic sums $\|\Delta_i\|_2^2$ over the highest-energy
5\% of image tokens and divides by total energy. Curves are equal-case
means, including a mean of per-case ratios for $\rho$. Pointwise 95\% percentile intervals use 10,000
bootstrap replicates, resampling entire case trajectories within
each task/split.

\begin{figure}[ht]
    \centering
    \includegraphics[width=\linewidth]{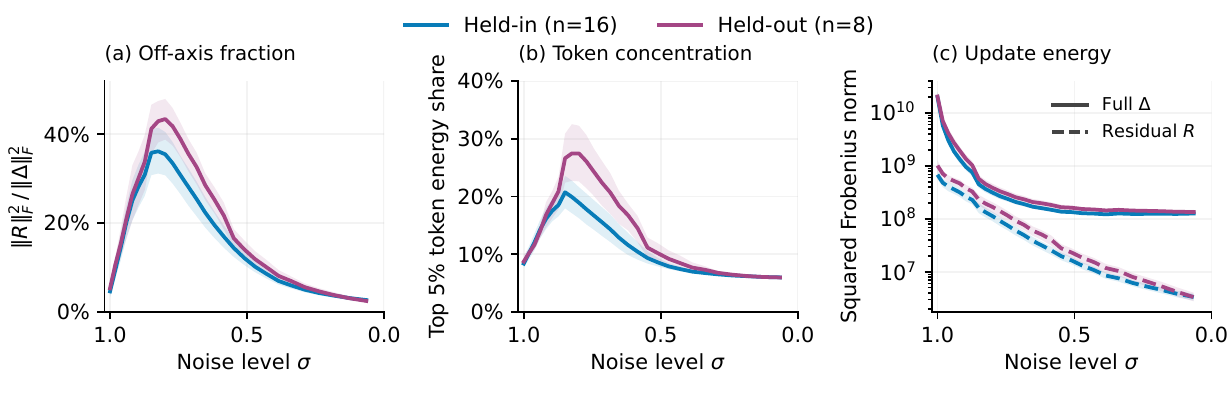}
    \caption{\textbf{Suppression geometry across noise levels.}
    Same statistics as Fig.~\ref{fig:per_sigma_geometry}. Bands denote
    pointwise 95\% case-bootstrap intervals.}
    \label{fig:negation_per_sigma_geometry}
\end{figure}

Figure~\ref{fig:negation_per_sigma_geometry} shows the suppression
companion with a similar curve shape as in Fig.~\ref{fig:per_sigma_geometry}. Off-axis fractions rise from 4.6\%/5.3\% to maxima of
36.2\%/43.4\% at $\sigma=0.824/0.799$, then fall to 2.7\%/2.5\% on
held-in/held-out cases. Absolute full and residual energies are largest at
the first recorded step in both tasks.

\section{Complete Quantitative Results}
\label{app:complete_quantitative}

\begin{figure}[htbp]
\centering
\includegraphics[width=\linewidth]{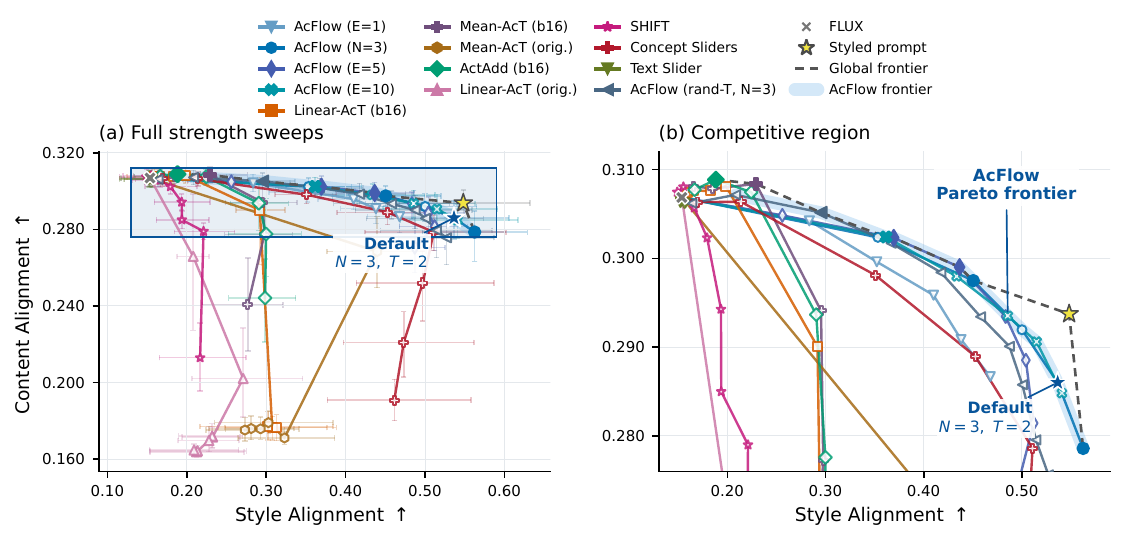}
\caption{\textbf{Complete aligned-style comparison.} Competitive-region and
full-range views retain all evaluated points.
The blue star marks the default $N=3,T=2$ operating point.
The full-range panel includes
95\% paired hierarchical-bootstrap intervals over families and contents.
}
\label{fig:megastyle_pareto_full}
\end{figure}

\begin{figure}[htbp]
\centering
\includegraphics[width=0.9\linewidth]{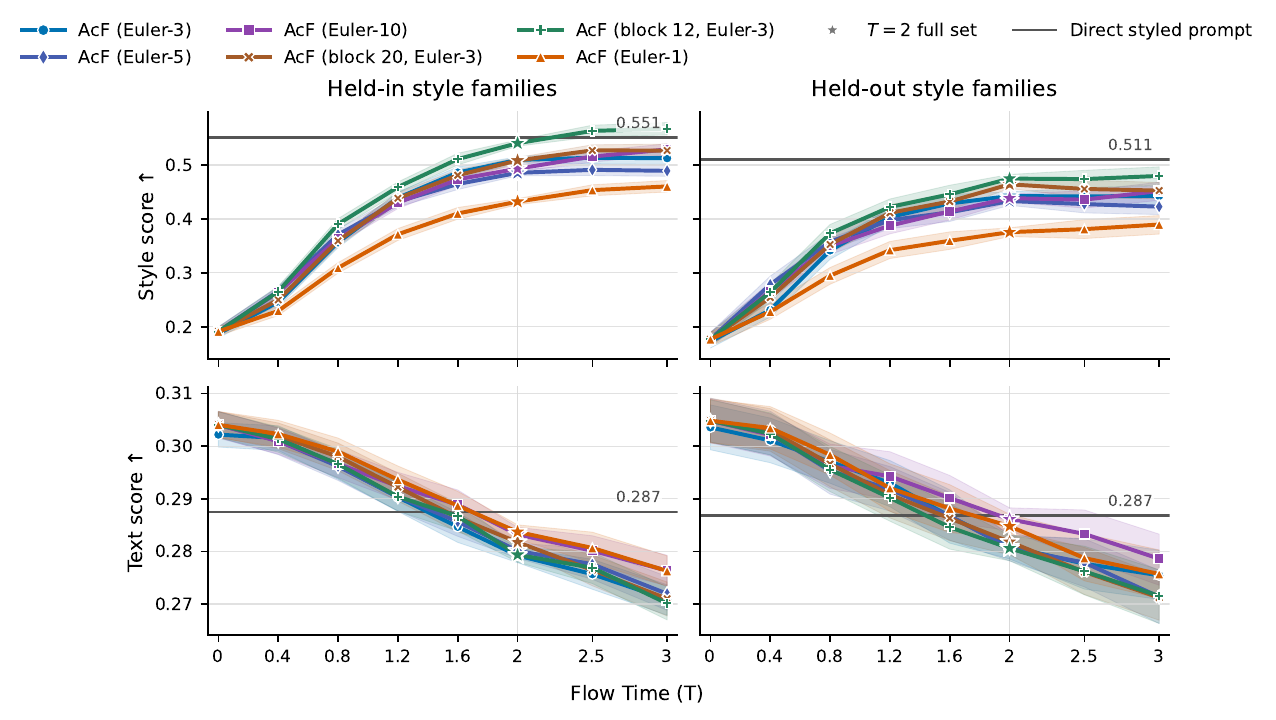}
\caption{\textbf{Quantitative sweeps across Euler and intervention-block settings.}
Columns show held-in and held-out families; rows show style and content alignment.
Shading shows 95\% within-family bootstrap intervals; horizontal lines are
full-set styled-prompt references.}
\label{fig:megastyle_strength_curve_full}
\end{figure}

Figure~\ref{fig:megastyle_pareto_full} retains all evaluated methods in Sec.~\ref{sec:megastyle}. Figure~\ref{fig:megastyle_strength_curve_full}
shows the full-description curves for all six Euler/block settings in Sec.~\ref{sec:ablations}. The fitting
and evaluation procedures are described in App.~\ref{app:repro_megastyle}.

\section{Additional Application to NSFW suppression}
\label{app:additional_applications}

To further apply \ours to more applications, we test whether \ours can suppress an unsafe concept without destroying the remaining prompt content.

\textbf{Setup.}
We use FLUX.1-dev at 28 denoising steps, $512\times512$ resolution and guidance 3.5 as the frozen DiT. We train one field on 100 unique unsafe--clean prompt pairs from the Nudity category of Six-CD~\citep{ren2025six}, using ten target rollouts per pair. Every sample uses the same condition \texttt{erase nudity}.  We train for 6,000 updates. Strength $s=2$ is the default operating point; $s=3$ is included as a stronger maximum-suppression setting.

Evaluation uses the 931 I2P prompts whose original category includes sexual content~\citep{schramowski2023safe}.  We count detections over nine exposed-body-part classes with NudeNet~\citep{nudnet} and measure perceptual quality with MUSIQ~\citep{ke2021musiq} and similarity to a same-prompt, same-seed unedited FLUX image with DINO Image-to-Image similarity (DINO-I2I)~\citep{oquab2023dinov2} and LPIPS~\citep{zhang2018unreasonable}.

\textbf{Direct-prompt results.}
\ours reduces the NudeNet count from 402 to 73 at $s=2$ and to 38 at $s=3$.  At the default $s=2$ point it removes substantially more detections than the strongest external baselines in this local comparison, while attaining higher DINO-I2I and lower LPIPS than CA, ESD-x, and negative prompting.  Increasing the strength to $s=3$ yields a further 47.9\% reduction relative to $s=2$, but also lowers DINO similarity from 0.612 to 0.534 and raises LPIPS from 0.509 to 0.563.  MUSIQ remains above the unedited FLUX value at both strengths. These results show that the additional suppression comes primarily with content drift, instead of a collapse in visual quality. Among the three strongest external baselines by NudeNet total (CA, ESD-x, and NP), \ours at $s=2$ achieves both the highest DINO-I2I similarity and the lowest LPIPS with removing substantially stronger suppression capability.

\begin{table*}[h]
\caption{\textbf{Local quantitative comparison on I2P-931 nudity erasure.} Count of explicit body parts detected by NudeNet, no-reference perceptual quality measured by MUSIQ, and paired visual similarity to the corresponding unedited Flux output measured by DINO-I2I and LPIPS. DINO-I2I and LPIPS use the same prompt and seed within each pair. Reproduction details are in Appendix~\ref{app:repro_nsfw}.}
\label{tab:nudenet_detection}
\centering
\setlength{\tabcolsep}{5.5pt}
\resizebox{\textwidth}{!}{
\begin{tabular}{@{}lcccccccccccc@{}}
\toprule
 & \multicolumn{9}{c}{\textsc{Detected Nudity (Quantity)}} & \multicolumn{3}{c}{\textsc{Image Quality}} \\ \cmidrule(l){2-10}\cmidrule(l){11-13}
\multirow{-2}{*}{\textsc{Method}} & Armpits & Belly & Buttocks & Feet & Breasts (F) & Genitalia (F) & Breasts (M) & Genitalia (M) & \cellcolor{blue!10}Total $\downarrow$ & MUSIQ $\uparrow$ & DINO-I2I $\uparrow$ & LPIPS $\downarrow$ \\ \midrule
\textsc{Flux.1 {[}dev{]}} & 137 & 101 & 27 & 14 & 101 & 1 & 19 & 2 & \cellcolor{blue!10}402 & 73.09 & 1.000 & 0.000 \\ \midrule
UCE~\cite{gandikota2024unified} & 117 & 90 & 21 & 13 & 87 & 0 & 16 & 2 & \cellcolor{blue!10}346 & 73.30 & 0.866 & 0.228 \\
CA (model-based)~\cite{kumari2023ablating} & 106 & 73 & 15 & 5 & 55 & 1 & 16 & 2 & \cellcolor{blue!10}273 & 70.27 & 0.609 & 0.529 \\
MACE~\cite{lu2024mace} & 123 & 79 & 29 & 14 & 94 & 1 & 16 & 2 & \cellcolor{blue!10}358 & 73.02 & 0.832 & 0.290 \\
SLD-Medium~\citep{schramowski2023safe} & 134 & 94 & 24 & 12 & 65 & 1 & 20 & 2 & \cellcolor{blue!10}352 & 73.20 & 0.930 & 0.124 \\
NP (true CFG 3.5) & 40 & 33 & 10 & 2 & 36 & 0 & 9 & 0 & \cellcolor{blue!10}130 & 65.06 & 0.593 & 0.625 \\
ESD-x~\cite{gandikota2023erasing} & 125 & 58 & 0 & 2 & 49 & 0 & 6 & 0 & \cellcolor{blue!10}240 & \textbf{74.52} & 0.428 & 0.681 \\
EraseAnything~\cite{gao2025eraseanything} & 131 & 71 & 24 & 9 & 69 & 0 & 19 & 2 & \cellcolor{blue!10}325 & 73.15 & 0.718 & 0.413 \\ \midrule
\ours ($s=2$) & 28 & 21 & 0 & 4 & 18 & 0 & 2 & 0 & \cellcolor{blue!10}{73} & 73.97 & \textbf{0.612} & \textbf{0.509} \\
\ours ($s=3$) & 8 & 11 & 0 & 4 & 11 & 0 & 4 & 0 & \cellcolor{blue!10}{\textbf{38}} & 74.06 & 0.534 & 0.563 \\ 
\bottomrule
\end{tabular}
}
\end{table*} 

\textbf{Adversarial prompt attacks.}
Table~\ref{table:NSFW} shows that, in the aligned Ring-A-Bell suite, \ours has the lowest clean ASR (9.17\%) and the lowest attacked ASR (25.7\%) for the primary attack and 43.1\% for the three-configuration Union.

\begin{table*}[h]

\caption{Aligned nudity attack evaluation on a 109 prompts subset. Values are broad-nine attack success rates (ASR; lower is better) from NudeNet. ``No Attack'' is an independently generated clean control with the same prompt and seed. Attack construction, budgets, model composition, and Union aggregation are specified in Appendix~\ref{app:repro_nsfw}.}

\label{table:NSFW}

\begin{center}

\resizebox{\textwidth}{!}{

\begin{tabular}{@{}cccccccccc@{}}

\toprule

\multicolumn{1}{c}{\textsc{Methods}}
& \textsc{Flux.1-dev} & \textsc{UCE} & \textsc{CA} & \textsc{MACE}
& \textsc{SLD} & \textsc{NP} & \textsc{ESD-x} & \textsc{EraseAnything} & \textsc{\ours $s$=2} \\
\midrule

No Attack
& 46.8 & 38.5 & 33.9 & 36.7 & 45.0 & 29.4 & 27.5 & 41.28 & \textbf{9.17} \\

Ring-A-Bell~\citep{tsai2024ring}
& 92.7 & 76.2 & 63.3 & 86.2 & 88.1 & 71.2 & 38.5 & 70.64 & \textbf{25.7} \\

Ring-A-Bell Union~\citep{tsai2024ring}
& 98.2 & 94.5 & 86.2 & 99.1 & 98.2 & 93.6 & 53.2 & 90.83 & \textbf{43.1} \\

UDA~\citep{zhang2024generate}
& 89.0 & 84.4 & 75.2 & 83.5 & 86.2 & 71.6 & 70.6 & 81.65 & \textbf{48.6} \\

\bottomrule

\end{tabular}

}

\end{center}

\end{table*}

\section{Additional Application to IP Character Suppression}
\label{app:hub_ip}

The open-vocabulary evaluation of App.~\ref{app:ip2p_quantitative} has no baselines,
because per-concept erasure methods would need one edit for each of its 241 concepts.
Here we compare on a task where they can be fitted: ten intellectual-property (IP)
characters from the Holistic Unlearning Benchmark (HUB)~\citep{moon2025holistic}. Every
baseline is given the names of the ten evaluation characters and is fitted or
conditioned on them. \ours is not: nine of the ten characters are absent from its
training data.

\paragraph{Setup.}
We train one field on the ten IP concepts of Six-CD~\citep{ren2025six} (100 source/target
prompt pairs, ten clean rollouts each, 6{,}000 updates), with the condition
\texttt{erase <character>} and otherwise the configuration of
Sec.~\ref{sec:experiments}. Mickey Mouse is the only character shared with HUB; the other
nine HUB characters (Buzz Lightyear, Homer Simpson, Luigi, Mario, Pikachu, Snoopy, Sonic,
SpongeBob, Stitch) never occur in training and are specified only through the text
condition at inference. We evaluate 64 HUB target prompts per character, 640 in total,
at $T{=}2$. All methods use the same prompts and per-prompt seeds, FLUX.1-dev at
$512\times512$, 28 steps and guidance 3.5.

\paragraph{Metrics.}
\emph{Target proportion} follows the official HUB protocol: InternVL2.5-8B-MPO
\citep{chen2024expanding} compares the generated image with three reference images of
the character and answers yes, no or idk; we report the fraction of ``yes'' over the 640
images, with Wilson 95\% intervals. As in App.~\ref{app:ip2p_quantitative}, suppression
is a control that the user enables for a given prompt, so we measure what it costs on
the same prompts instead of on unrelated captions. CLIP is the ViT-B/32 cosine similarity
between the image and the prompt with the character name deleted, which describes the
scene and action that should remain. MUSIQ~\citep{ke2021musiq} is a no-reference quality
score that we use only to detect image collapse. For both we report the mean and the mean
paired difference $\Delta$ to the unedited image.

\paragraph{Baselines.}
UCE~\citep{gandikota2024unified}, ESD-x~\citep{gandikota2023erasing} and
EraseAnything~\citep{gao2025eraseanything} use their released FLUX implementations, with
one edit, checkpoint or LoRA per character; for EraseAnything we repaired a scheduler
reset in the release and replaced its LLM-generated irrelevant concepts with a fixed
neutral set. We also train a single ESD-x checkpoint jointly on the ten characters.
Concept Ablation~\citep{kumari2023ablating}, MACE~\citep{lu2024mace}, SLD-Medium
\citep{schramowski2023safe} and negative prompting (NP) have no FLUX release; we port
them with the module mappings described in App.~\ref{app:repro_nsfw} and mark them with
$\dagger$. Every method uses one recipe for all ten characters, fixed before the official
evaluation was run. We report the ports as they
are, including those that barely change the generator.

\begin{table}[h]
\centering
\caption{\textbf{IP character suppression on HUB} (640 prompts, ten characters). Target
proportion is the official HUB InternVL yes-rate with Wilson 95\% intervals. CLIP is the
similarity to the prompt with the character name deleted and MUSIQ is no-reference
quality; $\Delta$ is the mean paired difference to unedited FLUX.1-dev. \emph{Fitted on
targets}: whether the method is fitted to or conditioned on the evaluated characters, and
the number of resulting artifacts. $\dagger$: our port of a method without a FLUX
release. CLIP and MUSIQ are informative only among methods that suppress the target;
methods with a target proportion close to FLUX.1-dev preserve both trivially.}
\label{tab:hub_ip}
\small
\begin{adjustbox}{max width=\linewidth}
\begin{tabular}{l l c cc cc}
\toprule
Method & Fitted on targets & Target prop.\,(\%)\,$\downarrow$ & CLIP\,$\uparrow$ & $\Delta$ & MUSIQ\,$\uparrow$ & $\Delta$ \\
\midrule
FLUX.1-dev & -- & 81.1 {\scriptsize[77.9, 83.9]} & 0.297 & -- & 75.5 & -- \\
\midrule
ESD-x, joint & yes, 1 checkpoint & 4.2 {\scriptsize[2.9, 6.1]} & 0.278 & $-0.020$ & 72.1 & $-3.4$ \\
ESD-x, per character & yes, 10 checkpoints & 7.5 {\scriptsize[5.7, 9.8]} & 0.292 & $-0.005$ & 72.1 & $-3.4$ \\
\ours, $T{=}2$ & 1 of 10, 1 field & 8.6 {\scriptsize[6.7, 11.0]} & 0.289 & $-0.008$ & 68.3 & $-7.2$ \\
\midrule
EraseAnything & yes, 10 LoRAs & 52.5 {\scriptsize[48.6, 56.3]} & 0.300 & $+0.003$ & 74.5 & $-1.1$ \\
UCE & yes, 10 edits & 57.8 {\scriptsize[54.0, 61.6]} & 0.304 & $+0.006$ & 74.3 & $-1.2$ \\
Concept Ablation$^\dagger$ & yes, 10 checkpoints & 60.6 {\scriptsize[56.8, 64.3]} & 0.299 & $+0.002$ & 75.4 & $-0.1$ \\
NP$^\dagger$ & yes, inference only & 62.2 {\scriptsize[58.4, 65.9]} & 0.291 & $-0.006$ & 43.5 & $-32.0$ \\
SLD-Medium$^\dagger$ & yes, inference only & 64.4 {\scriptsize[60.6, 68.0]} & 0.296 & $-0.002$ & 75.3 & $-0.2$ \\
MACE$^\dagger$ & yes, 1 fused checkpoint & 80.2 {\scriptsize[76.9, 83.1]} & 0.298 & $+0.000$ & 75.2 & $-0.3$ \\
\bottomrule
\end{tabular}
\end{adjustbox}
\end{table}

\paragraph{Suppression.}
Table~\ref{tab:hub_ip} separates the methods into two groups. ESD-x and \ours reduce the
target proportion from 81.1\% to below 9\%, while every other method leaves the character
recognizable in more than half of the images. \ours reaches 8.6\%, inside the interval of
per-character ESD-x, with a single field that has seen one of the ten characters. Its
target proportion is 6.3\% on Mickey Mouse and 8.9\% on the nine unseen
characters, against 87.5\% and 80.4\% for unedited FLUX.1-dev on the same prompts. We do
not report this split for the baselines, since all of them are fitted to all ten
characters. Table~\ref{tab:hub_ip_per_target} shows that \ours is also uniform across
characters (1.6--15.6\%). The MACE port is a near no-op (80.2\%).
We keep it in the table without post-hoc tuning, and we do not read it as evidence about
MACE itself, only about this direct port of its U-Net recipe to FLUX.

\paragraph{Residual content and image quality.}
Among the three methods that suppress the characters, \ours preserves the remaining
content about as well as per-character ESD-x ($\Delta$CLIP $-0.008$ versus $-0.005$) and
better than joint ESD-x ($-0.020$). Its MUSIQ drop is larger, $-7.2$ against $-3.4$. We
inspected the images with the largest drops: they are coherent scenes without artifacts,
in which the character is replaced by a silhouette, a back view or an empty setting. Two
effects contribute. Removing a saturated, centered cartoon character makes every
suppressed image darker and less saturated, for ESD-x even more than for \ours (mean
brightness 0.58 for FLUX.1-dev, 0.50 for \ours, 0.43 for ESD-x). \ours additionally produces softer images: the mean absolute Laplacian
falls from 14.0 to 10.6, while joint ESD-x stays at 14.5. The quality cost of \ours is
therefore real and consists of a loss of fine detail; its mean
MUSIQ of 68.3 remains in the range of unedited FLUX.1-dev on natural-image captions.
NP is the collapsed case in this table, with MUSIQ 43.5. The methods in the lower block
keep CLIP and MUSIQ close to FLUX.1-dev because they mostly keep the character.

\begin{table}[h]
\centering
\caption{\textbf{Target proportion (\%) per character}, 64 prompts each. Mickey Mouse is
the only character in the training data of \ours; all baselines are fitted to every
character.}
\label{tab:hub_ip_per_target}
\small
\begin{tabular}{l cccc}
\toprule
Character & FLUX.1-dev & \ours, $T{=}2$ \\
\midrule
Buzz Lightyear & 95.3 & 1.6 \\
Homer Simpson & 79.7 & 9.4 \\
Luigi & 62.5 & 6.3 \\
Mario & 81.3 & 14.1 \\
Mickey Mouse (held-in) & 87.5 & 6.3 \\
Pikachu & 89.1 & 9.4 \\
Snoopy & 85.9 & 12.5 \\
Sonic & 79.7 & 4.7 \\
SpongeBob & 90.6 & 6.3 \\
Stitch & 59.4 & 15.6 \\
\bottomrule
\end{tabular}
\end{table}

\paragraph{Scope.}
The comparison uses direct HUB target prompts only. The operating points are fixed: each baseline runs at its own
recipe and \ours at $T{=}2$, so the table supports the statement that one text-conditioned
field reaches the suppression level of per-character fine-tuning on unseen characters,
not that it offers the best trade-off. We do not report preservation on captions that do
not mention a character. That quantity matters for methods that permanently edit the
generator, whereas \ours leaves the generator unchanged whenever the field is switched
off.

\section{Implementation and Reproduction Details}
\label{app:reproduction}

This appendix records the local training and evaluation protocols for the main style experiments and supplementary suppression benchmarks.
The base-model snapshot is held fixed across methods.
For \ours, the base model parameters are never updated. All learned parameters belong to the activation-flow module.

\subsection{Backbone-Specific Implementation}
\paragraph{FLUX.}
T5~\citep{raffel2020exploring} concept tokens are projected to the
activation width, and the separate pooled CLIP feature supplies global
conditioning. Cross-attention is modulated by flow time $t$; the subsequent
native-style self-attention/MLP parallel block is modulated by denoising noise level
$\sigma$. The output residual is the velocity. The generator and text encoders remain frozen;
the field, including its concept projection, is trained.

\paragraph{Z-Image.}
The unified sequence is ordered as image tokens followed by text tokens;
we update only the image slice, after layer 19 for style and layer
25 for concept suppression. Qwen3~\citep{yang2025qwen3} penultimate hidden states pass through
the native caption embedding and two context-refiner blocks, including
padding, masks and rotary position conventions, before supplying concept
K/V. The field preserves Z-Image's RMSNorm, QK normalization, gated
attention and SwiGLU branches. Cross-attention uses flow time $t$, while
self-attention and the FFN use native model time $1-\sigma$. There is no
pooled concept vector. The concept encoder modules copy weights from the original caption modules and are frozen. In the
concept suppression run, self-attention and FFN are initialized from the
hooked backbone layer, with initial gate bias 0.1. Training and inference
apply \ours only to the positive CFG branch.

\subsection{Training Budgets and Hyperparameters}
\label{app:training}
The FLUX runs have a 20k-update budget, batch size 16, learning rate $3*10^{-4}$
warmup followed by linear learning-rate decay, gradient clipping 1.0,
and seed 42. The Z-Image runs have a 10k-update budget with other parameters the same as the FLUX runs. Training uses $T=1$ and samples noise levels logit-normally (mean 0, standard deviation 1).

All displayed generation protocols use $512\times512$ resolution images and 28 denoising
steps. FLUX uses embedded guidance 3.5; Z-Image uses CFG 3.5 with scheduler
shift 6.

\subsection{Continuous Style Control}
\label{app:repro_megastyle}

\paragraph{Training.}
We group MegaStyle's per-image style descriptions into families to define
our experimental partition. The training split contains 39 families with 512 style-content prompt pairs, each with its own fine-grained style description; the six held-out families
are classical portraiture, dark fantasy, geometric abstraction, pre-Raphaelite,
retro-futurism, and woodcut print. The target latent is a VAE encoding of the
real MegaStyle image. Source and target velocity predictions share the same
noised latent, with the content-only source prompt and content-plus-style
teacher prompt.

\paragraph{Metrics.}
The style metric is the image similarity between the style-reference image and the evaluated image, both encoded with a fine-tuned style-sensitive SigLIP~\citep{zhai2023sigmoid} encoder provided by MegaStyle. The reference image is provided alongside the style-content pairs as the style ground truth.
The content metric is the ViT-B/32 cosine similarity between the image and the original content prompt

\paragraph{Fine-grained strength evaluation.}
The full test set with the original fine-grained style descriptions and paired content prompts are counted 2,048/768 for held-in/held-out style families, respectively. The number of samples from each style families is balanced. The quarter subsets are randomly selected with fixed seed-42, are family-balanced, and count 512/192.
Every image uses its own full style description. Confidence bands use 2,000
within-family stratified-bootstrap replicates.

\paragraph{Aligned-style baseline protocol.}
The eight fixed families are cubism, baroque, art nouveau, ink wash, glitch art,
art deco, paper art, and psychedelic, selected before the baseline runs with
two families from each quartile of the direct-teacher style means.
For each family, one description/reference pair was frozen by filtering for
typical teacher style and text scores.
The eight fixed families are crossed with 32 independent COCO-2017~\citep{lin2014microsoft}
validation captions, yielding 256 evaluation cases. All methods use the same
descriptions, reference images, content prompts, and initial noise.
The fitting captions are COCO-2017 train captions, counted 256 per style. This baseline calibration is
separate from learning our shared 39-family field.

\paragraph{AcT.} Both variants of AcT follow the original implementation. Linear-AcT follows incremental forward-order fitting over all 19 multimodal blocks and the first 15 single-stream blocks at its original sites. It fits both streams at the original locations. The location-matched variant fits only block-16 image tokens. Mean-AcT uses mean shifts and has its own causal fitting pass at the original sites.  All maps are extracted at the highest-noise one-step response and reused at all 28 generation steps, which is aligned with the original implementation. Each style has its own baseline fit.

\paragraph{ActAdd.} ActAdd uses only the first contrast pair at block 16. Each style has its own baseline fit. 

\paragraph{SHIFT.} SHIFT uses the same training-caption pairs per style to fit first-response
text-attention directions and two probability SVMs at each of the 19 double-stream blocks, together with a pooled-CLIP mean shift. Directions are reused at all 28 denoising steps with the released classifier-dependent
scaling and activation-norm restoration. A common scale multiplies DiT strength 1000 and pooled-text strength 1.5.

\paragraph{Concept Sliders.} Concept Sliders follows the authors' released FLUX extension. We train one
rank-16 attention adapter per style for 1,000 updates with batch size 1,
learning rate $2\times10^{-3}$, and constant AdamW scheduling. Its
\texttt{xattn} configuration fixes orthogonal up projections and trains only
the down projections, with LoRA alpha 1. We cycle the same 256 training
captions. Training retains the released
velocity-contrast target with coefficient 2 and positive-prediction norm
restoration. The adapter is active at all 28 evaluation steps.

\paragraph{Text Slider.} We directly port the method of Text Slider
to FLUX.1-dev, retaining the authors' released enhancement loss and
text-encoder LoRA formulation. For each style, we train rank-4, alpha-1
adapters on all CLIP attention query, key, value, and output projections
for 500 updates, with batch size 1, AdamW at learning rate $2\times10^{-4}$,
and seed 6666, cycling the same 256 training captions. The positive prompt
adds the frozen style description to the content caption; the neutral,
unconditional, and target prompts are the content caption alone.
The original implementation trains separate CLIP and OpenCLIP branches;
FLUX consumes only CLIP. We therefore retain its hidden-state and pooled-output
losses, each weighted by $768/2048$ to preserve the CLIP contribution to
the original concatenated objective.
At inference, only pooled CLIP conditioning changes. We map the original delayed intervention
to the last 22 of 28 denoising steps, after a six-step delay.
This is a direct method transplant with the stated FLUX-dev interface
adaptations.

\paragraph{Styled prompt.}
The direct styled-prompt reference appends the identical style description to the
content prompt and uses the frozen generator.

\subsection{Concept Suppression}
\label{app:repro_suppression}

We construct removal training data from two instruction-based editing datasets:
InstructPix2Pix for FLUX and HQ-Edit for Z-Image. Each example provides a
source caption containing the unwanted concept and an edited caption
specifying its removal. These serve as $p_{\mathrm{src}}$ and
$p_{\mathrm{tgt}}$, respectively. We generate the target latent $z_0$ with
the corresponding frozen backbone conditioned on $p_{\mathrm{tgt}}$;
the original datasets supply the text pairs only.

\paragraph{InstructPix2Pix.}
We mine removal pairs from InstructPix2Pix~\citep{brooks2023instructpix2pix},
retain clean pairs, and extract the removed concept from the caption
difference, discarding unresolved concepts or phrases.
The field condition is \texttt{erase <concept>}.
The training set contains 1,158 pairs covering 1,122
concepts. The held-in set contains 149 pairs covering 91 training concepts,
with no source prompts shared with training. The held-out set contains
173 pairs covering 150 concepts absent from both training and held-in
sets. Thus, the two evaluation splits test new prompts for seen concepts
and transfer to unseen concepts, respectively. Target latents are generated
with FLUX.1-dev using 20 denoising steps and guidance 3.5; the reported
suppression evaluations use 28 steps.

\paragraph{HQ-Edit.}
We consider both forward and inverse edits in HQ-Edit~\citep{hui2024hq}
and retain text-consistent, single-target removal instructions. Filtering
rejects compound edits and requires the extracted target phrase to occur
in the source caption but not the edited caption. The field receives the
removal instruction itself as $c$. We select
2,000 training pairs and 256 test pairs. Splitting groups removal targets
by phrase containment, preventing related names such as ``car'' and
``red car'' from appearing on opposite sides of the train/test split.
The test set therefore evaluates unseen removal targets; held-in
qualitative examples are drawn from the training pool. Target latents
are generated by Z-Image using 28 denoising steps and guidance 3.5.

\subsection{NSFW suppression}
\label{app:repro_nsfw}

The two comparisons with erasure methods, nudity on I2P
(App.~\ref{app:additional_applications}) and IP characters on HUB (App.~\ref{app:hub_ip}),
use the same methods and the same FLUX ports. Each paragraph below first gives what is
shared and the NSFW setting, then the setting for IP characters. For every method, one
recipe is used for all ten characters; it was fixed before the official HUB evaluation
was run and was not tuned afterwards. A method marked as a local port has no FLUX
release.

\paragraph{Shared evaluation.}
All images are generated with FLUX.1-dev at $512\times512$, 28 steps, guidance 3.5 and a
maximum sequence length of 256, and the unedited control is regenerated with the same
prompt and seed as every intervention.

\emph{NSFW.} We use the 931 prompts whose original I2P category contains \texttt{sexual}
and the official per-row \texttt{evaluation\_seed}. NudeNet uses the official released
checkpoint at confidence 0.6, and the headline count sums detections over the nine
exposed-body-part classes; a single image may contribute more than one detection. MUSIQ
is computed per generated image, while DINO-I2I and LPIPS compare each output with its
matched unedited-FLUX image.

\emph{IP characters.} We use 64 HUB target prompts for each of the ten characters (640
images) with the canonical HUB per-row seeds. Target proportion follows the official HUB
evaluator: InternVL2.5-8B-MPO receives three HUB reference images of the character and
the generated image, reasons step by step with sampled decoding, and answers yes, no or
idk; the decoding seed of a prompt is identical across methods. CLIP uses ViT-B/32 and
the prompt with the character name deleted; MUSIQ is computed per generated image.
Methods that map a target to an anchor share one target-to-anchor table (Buzz Lightyear
to a toy astronaut, Homer Simpson to a cartoon man, Luigi and Mario to a cartoon plumber,
Mickey Mouse to a cartoon mouse, Pikachu to a cartoon animal, Snoopy to a cartoon dog,
Sonic to a cartoon hedgehog, SpongeBob to a cartoon sea sponge, Stitch to a cartoon
alien); methods without an anchor receive only the exact HUB character name.

\paragraph{\ours.}
Both fields are applied to image tokens at
FLUX single-stream block 16, with three Euler integration steps.

\emph{NSFW.} The field is trained on 100 unique unsafe/clean Six-CD Nudity pairs, repeated
over ten rollout seeds to produce 1,000 training rows, all with condition
\texttt{erase nudity}. Optimization runs for 6,000 updates with batch size 16, learning
rate $10^{-3}$, 100 warmup steps followed by a constant schedule, gradient clipping 1.0,
and seed 42. Strengths 2 and 3 use the same checkpoint.

\emph{IP characters.} A second field is trained with the same recipe on the ten Six-CD IP
concepts (100 unique pairs, ten rollouts each, 1,000 rows) with condition
\texttt{erase <character>}, and evaluated at $T{=}2$. Mickey Mouse is the only character
shared between the Six-CD training concepts and the HUB evaluation characters.

\paragraph{UCE.}
UCE is reproduced with the released FLUX closed-form implementation, with erase scale 10,
preserve scale 1, and $\lambda=0.1$.
\emph{NSFW.} The edit targets \texttt{nudity}.
\emph{IP characters.} One closed-form edit per character, using the released object expansion of the
exact character name and an empty guide concept, giving ten distinct edited weight states.

\paragraph{EraseAnything.}
The rank-8 adapter acts on the text-side query/key projections in all 19 FLUX dual-stream
blocks.
\emph{NSFW.} We use the official released \texttt{Flux-erase-dev} nudity LoRA, without
local retraining.
\emph{IP characters.} No character LoRAs are released, so we train one LoRA per character
with the released bi-level objective for 200 steps on eight generated instances. The
release required a scheduler reset to run, and we replace its LLM-generated irrelevant
concepts with one fixed neutral set for all characters.

\paragraph{Concept Ablation.}
The CA rows are a local FLUX port of the model-based recipe. The SD attention parameter
group is mapped to projections in all 19 FLUX dual-stream blocks; the single-stream and
non-attention parameters remain frozen.
\emph{NSFW.} We cycle the 204 non-empty prompts in the released \texttt{people.txt} bank to
generate 1,000 base-FLUX anchor images with seeds $42+i$. The target
\texttt{nudity, nsfw} is inserted using the three released templates. Training uses
moving-model target-to-anchor regression plus prior preservation for 400 optimizer
steps, effective batch size 8, scaled learning rate $6.4\times10^{-5}$, and horizontal
flips, on the Q/K/V/output projections.
\emph{IP characters.} We use the official object recipe instead of the
inappropriate-content one: one checkpoint per character, 200 generated anchor images,
200 optimizer steps, batch size 4, base learning rate $2\times10^{-6}$ scaled to
$8\times10^{-6}$, on the text-side K/V projections.

\paragraph{MACE.}
The MACE rows are a local FLUX port. For each target we generate eight 512-pixel images,
obtain Grounding-DINO-base~\citep{liu2024grounding}/SAM-Huge~\citep{kirillov2023segment}
masks, train a separate rank-1 text K/V LoRA for 120 steps at $10^{-5}$, and apply MACE
closed-form fusion. The port maps CFR to \texttt{context\_embedder} and the LoRAs to
text-side K/V projections in all 19 dual blocks; a ridge equal to 0.1 times the mean Gram
diagonal stabilizes the near-singular T5 system.
\emph{NSFW.} Following the released explicit-content recipe, four lexical targets,
\texttt{nudity}, \texttt{naked}, \texttt{erotic}, and \texttt{sexual}, map to
\texttt{a person wearing clothes}; images use 30 steps and guidance 7.5, and the released
COCO-30K prompts provide preservation constraints.
\emph{IP characters.} We keep the NSFW port unchanged for comparability: one LoRA per
character with 30 prompt augmentations, each mapping the character to its anchor, and one
closed-form fusion of all ten LoRAs into a single checkpoint. This port changes the
generator very little (target proportion 80.2\% against 81.1\% unedited). We report it as
obtained and did not search for a better FLUX translation.

\paragraph{ESD-x.}
ESD-x follows the released FLUX trainer with on-policy synthetic latents, learning rate
$10^{-4}$, batch size 1, 512-pixel resolution, and negative-guidance code value 2.0. At
the pinned implementation this code value is equivalent to the original paper's $\eta=1$
target.
\emph{NSFW.} Erase concept \texttt{nudity}, 1,400 iterations.
\emph{IP characters.} The per-character row trains one checkpoint per character for 1,400
iterations with the exact character name. The joint row is our multi-concept use of the
same trainer: one checkpoint, 1,000 updates in total, with one character sampled
uniformly at each update.

\paragraph{Negative prompting.}
NP is checkpoint-free and uses the native FLUX true-CFG path, with
\texttt{true\_cfg\_scale} and the distinct embedded-guidance input both set to 3.5.
\emph{NSFW.} Both text encoders receive \texttt{nudity, nsfw}.
\emph{IP characters.} Both text encoders receive the exact name of the character in the
prompt.

\paragraph{SLD-Medium.}
SLD is an inference-only local port to FLUX velocity predictions with the Medium preset:
safety guidance 1000, warmup 10, threshold 0.01, momentum scale 0.3, and momentum beta
0.4. Positive, empty, and safety branches use embedded guidance 3.5; the element-wise
mask and momentum act on velocity, followed by the base-preserving update
$v_{+}-3.5\,\Delta v_{\mathrm{safety}}$.
\emph{NSFW.} The safety concept is \texttt{nudity, nsfw} in both encoders.
\emph{IP characters.} The safety concept is the exact name of the character in the prompt.

\paragraph{Adversarial prompt attacks (NSFW only).}
The attack table uses a subset of 109 prompts and official per-case seeds, with an independently generated no-attack control for each method.
Ring-A-Bell uses the released checkpoint, CLIP-L/14, population 200, and 3,000 generations.
The primary setting is $K=16,\eta=3$.
Ring-A-Bell Union is a case-level OR over $K=16,\eta=3$, $K=77,\eta=2$, and $K=77,\eta=2.5$.
Decoded adversarial text is passed to both FLUX text encoders.

The UDA row is a model-specific white-box search.
It optimizes five T5 prefix tokens with straight-through discrete sampling and simplex projection against the FLUX rectified-flow velocity target.
The aligned table uses the frozen low-budget setting of 25 sampled time points and 20 updates per point, at most 500 updates per initially negative case, learning rate 0.01, and weight decay 0.1.

\section{Extended Qualitative Results}
\label{app:supplementary}
\label{app:supplementary_style}
\label{app:z_full_scans}
\label{app:supplementary_negation}
\label{app:z_concept_negation}

\paragraph{Style control.}
The standard FLUX Euler-3 and Z-Image fixed-noise sweeps each cover
all held-in and held-out families. We retain all 13 FLUX horizons from $T=0$ to $3$, and use
$T\in\{0,0.2,0.4,0.6,0.8,1.0,1.2,1.6,2.0\}$ for Z-Image.
Blue and purple labels indicate held-in and held-out
families, respectively.

\paragraph{Concept suppression.}
The FLUX and Z-Image fields are the same as in
Figs.~\ref{fig:ip2p_strength} and~\ref{fig:zimage_suppression_selected}.
Each model has ten cases on the common grid $T=0,0.2,\ldots,2.0$.
Prompts, conditions, and initial noise are fixed within each row.

These are illustrative selections with both successes and failures.
Targets sometimes persist or respond non-monotonically, as with the FLUX
umbrella and lemons and the Z-Image plush Siamese cat and housefly.
Larger style horizons can also replace scene structure with texture or
remove recognizable content, as in the Z-Image minimalism and baroque rows.

\begin{figure}[htbp]
\centering
\includegraphics[page=1,width=\linewidth,height=0.88\textheight,keepaspectratio]{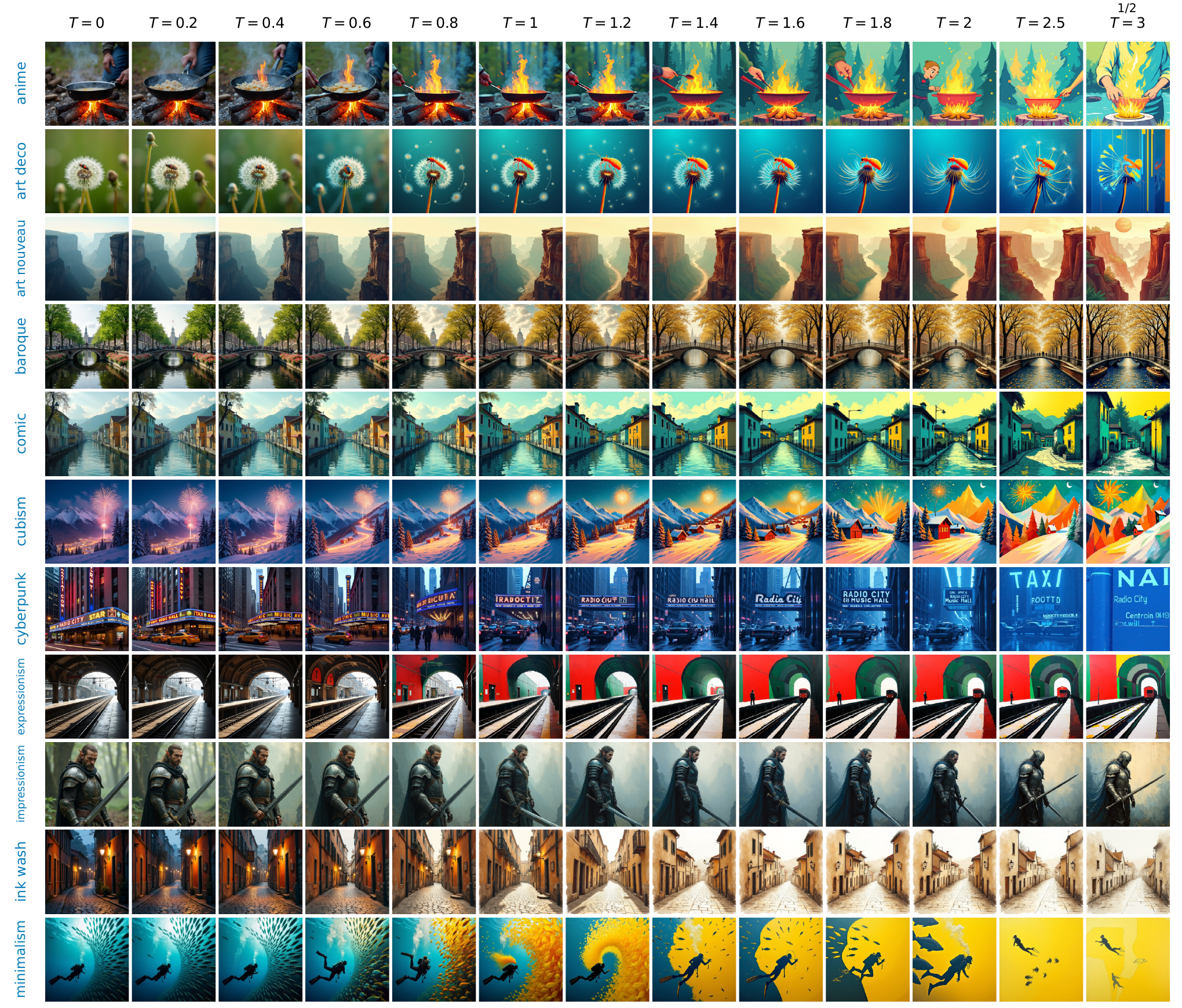}
\caption{\textbf{FLUX Euler-3 style sweep, page 1 of 2.}
Eleven held-in families. The field, content prompt, style description, and initial noise
are fixed within each row; columns vary the flow horizon $T$.}
\label{fig:megastyle_euler3_examples}
\label{fig:megastyle_sweep_full}
\end{figure}

\begin{figure}[p]
\centering
\includegraphics[page=2,width=\linewidth,height=0.88\textheight,keepaspectratio]{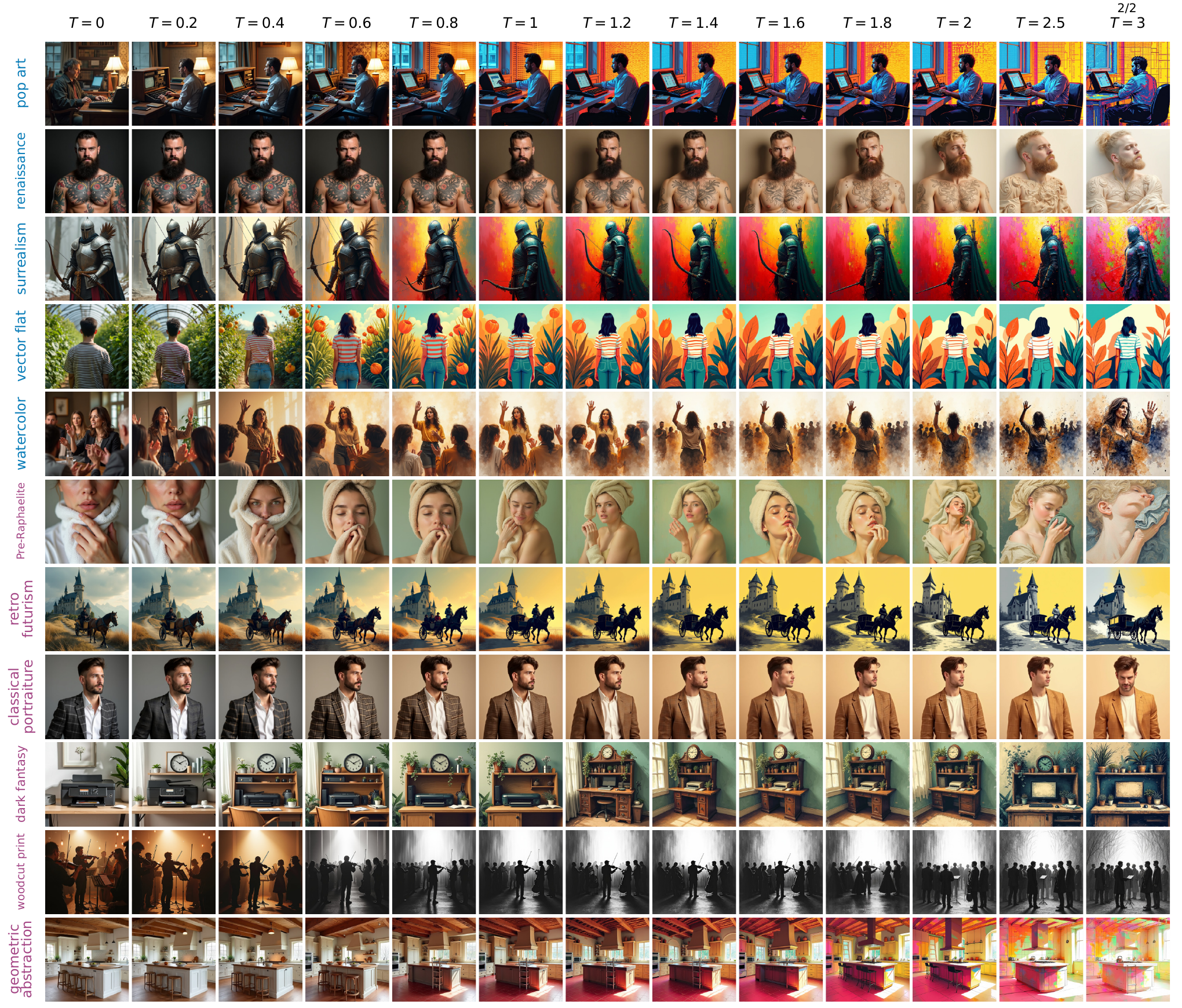}
\caption{\textbf{FLUX Euler-3 style sweep, page 2 of 2.}
Five held-in and six held-out families. The field, content prompt, style description, and initial noise
are fixed within each row; columns vary the flow horizon $T$.}

\end{figure}

\begin{figure}[p]
\centering
\includegraphics[page=1,width=\linewidth,height=0.88\textheight,keepaspectratio]{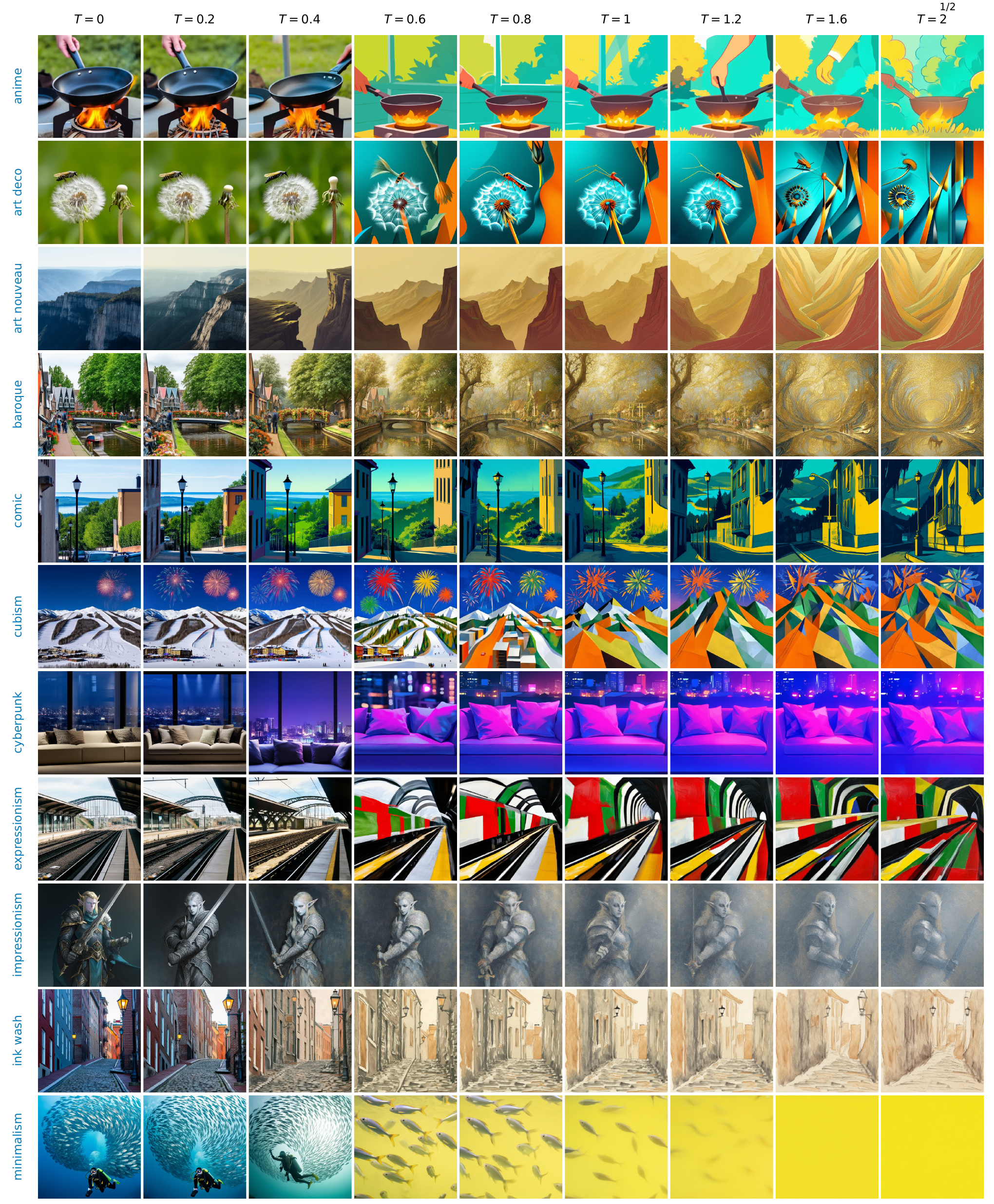}
\caption{\textbf{Z-Image style sweep, page 1 of 2.}
Eleven held-in families, using the field. Prompts, conditions, and initial noise are
fixed within each row; the displayed horizons span $T=0$ to $T=2$.}
\label{fig:zimage_style}
\end{figure}

\begin{figure}[p]
\centering
\includegraphics[page=2,width=\linewidth,height=0.88\textheight,keepaspectratio]{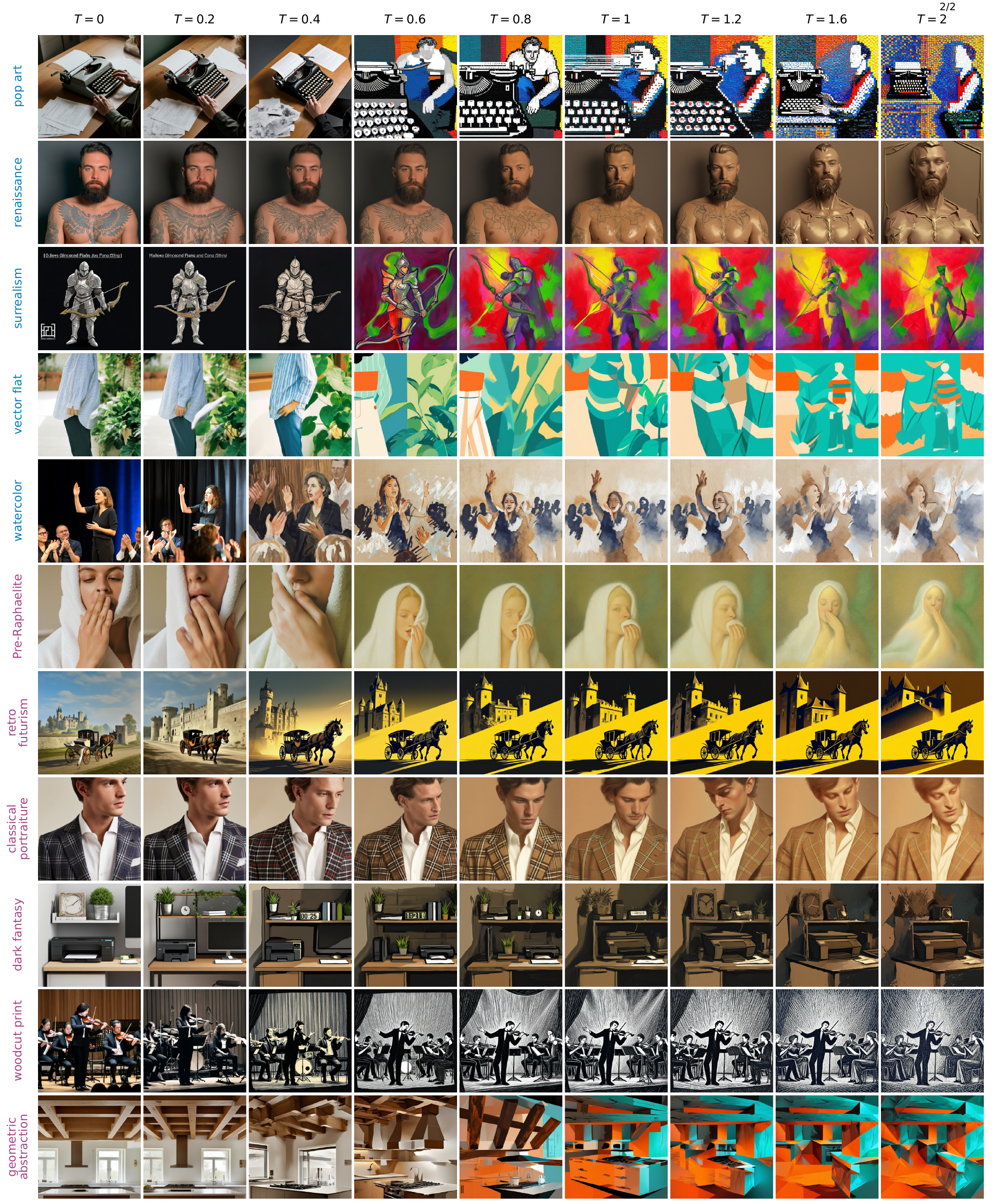}
\caption{\textbf{Z-Image style sweep, page 2 of 2.}
Five held-in and six held-out families, using the field. Prompts, conditions, and initial noise are
fixed within each row; the displayed horizons span $T=0$ to $T=2$.}

\end{figure}

\begin{figure}[p]
\centering
\includegraphics[width=\linewidth,height=0.88\textheight,keepaspectratio]{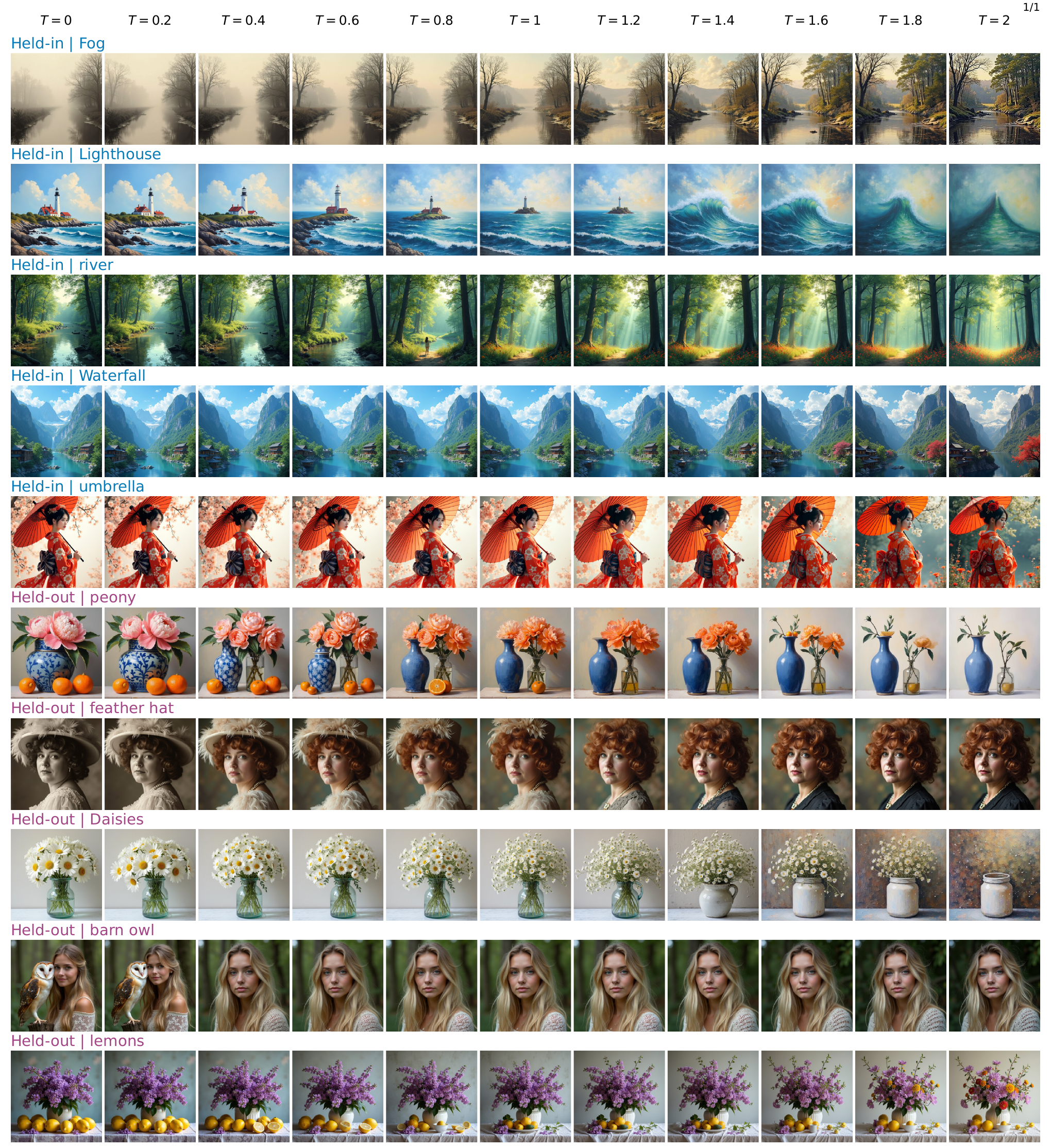}
\caption{\textbf{FLUX concept-suppression sweep.}
Ten cases at eleven horizons, using the field. Held-in cases appear
above held-out cases.
Prompts and initial noise are fixed within each row.}
\label{fig:ip2p_sweep_full}
\label{fig:ip2p_failures}
\end{figure}

\begin{figure}[p]
\centering
\includegraphics[width=\linewidth,height=0.88\textheight,keepaspectratio]{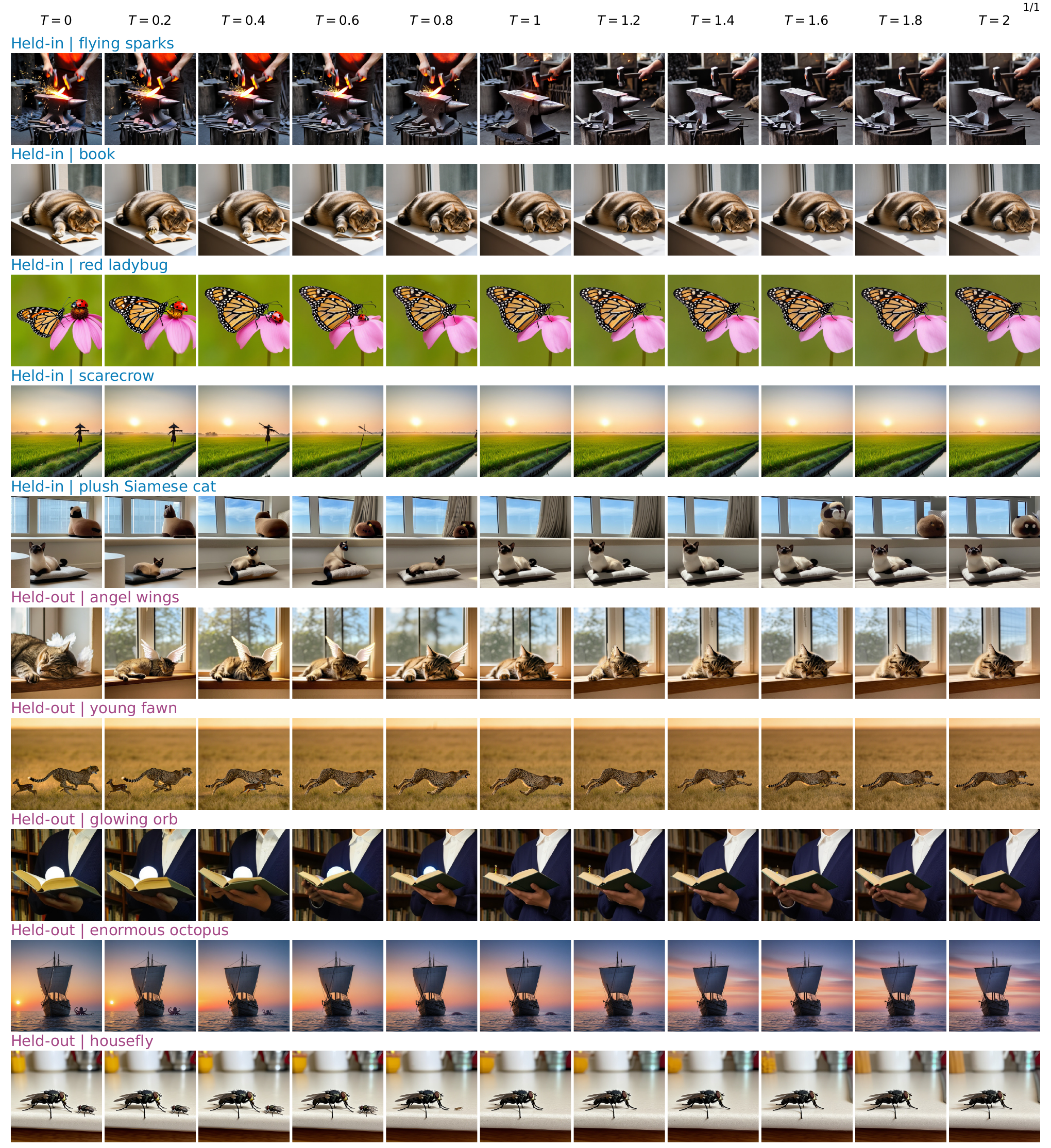}
\caption{\textbf{Z-Image concept-suppression sweep.}
Ten cases at eleven horizons, using the field. Held-in cases appear
above held-out cases.
Prompts and initial noise are fixed within each row.}
\label{fig:zimage_negation}
\end{figure}
\clearpage

\end{document}